\documentclass[10pt,journal,compsoc]{IEEEtran}

\usepackage{amsmath,graphicx}
\usepackage[caption=false]{subfig}
\usepackage[keeplastbox]{flushend}
\usepackage{multirow}
\usepackage{color}
\usepackage{multicol}
\usepackage{cite}

\begin{document}

\thispagestyle{empty}

\onecolumn
\textbf{IEEE Copyright Notice}

© 2020 IEEE. Personal use of this material is permitted. Permission from IEEE must be obtained for all other uses, in any current or future media, including reprinting/republishing this material for advertising or promotional purposes, creating new collective works, for resale or redistribution to servers or lists, or reuse of any copyrighted component of this work in other works.

\twocolumn

\clearpage
\pagenumbering{arabic} 

%
\title{Frequency learning for image classification}

\author{Jos\'{e} Augusto Stuchi,
        Levy Boccato
        and Romis Attux}


\IEEEtitleabstractindextext{%

\begin{abstract}
Machine learning applied to computer vision and signal processing is achieving results comparable to the human brain on specific tasks due to the great improvements brought by the deep neural networks (DNN). The majority of state-of-the-art architectures nowadays are DNN related, but only a few explore the frequency domain to extract useful information and improve the results, like in the image processing field. In this context, this paper presents a new approach for exploring the Fourier transform of the input images, which is composed of trainable frequency filters that boost discriminative components in the spectrum. Additionally, we propose a slicing procedure to allow the network to learn both global and local features from the frequency-domain representations of the image blocks. The proposed method proved to be competitive with respect to well-known DNN architectures in the selected experiments, with the advantage of being a simpler and lightweight model. This work also raises the discussion on how the state-of-the-art DNNs architectures can exploit not only spatial features, but also the frequency, in order to improve its performance when solving real world problems.


\end{abstract}

\begin{IEEEkeywords}
machine learning, Fourier analysis, image classification, deep learning, frequency filtering
\end{IEEEkeywords}}

\maketitle

\IEEEdisplaynontitleabstractindextext

%
\IEEEpeerreviewmaketitle


\IEEEraisesectionheading{\section{Introduction}\label{sec:introduction}}
\label{sec:intro}


Machine learning (ML) has impacted many applications in the last years, especially in computer vision and signal processing areas\cite{goodfellow2016deep}. In particular, the number of tasks successfully addressed with artificial neural networks (ANNs) and deep learning (DL) is exploding, since these methods grant more reliability to the machine to solve complex problems, such as medical diagnosis assistance and autonomous vehicles. 

Inspired on the brain plasticity, ANNs are capable of adapting their behavior and learning from the available data \cite{haykin2009neural}. The chosen neuron model, how they are distributed and connected, and what are the adaptable parameters basically define the network architecture. Given a particular architecture, the ANN can learn from data and build experimental knowledge, making it available while making predictions for new data \cite{haykin2009neural}.

The progress of the learning algorithms, the availability of large datasets and the possibility of using highly parallel hardware, with graphic processing units (GPU), greatly boosted the success of ANN/DL applications in the last decade. 
In this scenario, deep and more complex architectures started to be used to solve difficult problems in many different fields, being the state-of-the-art in many areas, such as speech and text processing, and computer vision\cite{lecun2015deep}. 

Traditionally, the effective use of machine learning techniques involved a pre-processing stage, in which the available raw data were transformed into proper features, usually with a reduced dimensionality when compared to the original data. This approach requires a careful design of the feature extractor by experts with high domain knowledge, in order that discriminative features are attained and, ultimately, the ML model reaches an adequate performance. Some well-known examples of handcrafted features are the mel-frequency cepstral coefficients (MFCC)\cite{davis1980comparison} for sound processing, and speeded up robust features (SURF)\cite{bay2006surf} for image processing. Both techniques demanded a long time and a highly specialized research team to be developed. 


On the other hand, the approach explored in deep learning is to let the model learn directly from the raw data. In this sense, the deep model indirectly creates efficient internal representations for the data while it is learning how to perform the desired task (e.g., to classify images). Additionally, the deep and layered structure of the model gives rise to a hierarchical representation, since the first layers tend to extract low level features, which are progressively combined into high level features by the subsequent layers. Given the high flexibility of the deep models, this perspective proved to be advantageous in several challenging real-world tasks, leading to superior results when compared with using generic handcrafted features. Among the most prominent deep models, the convolutional neural networks (ConvNets)\cite{goodfellow2016deep} stand out in the field of computer vision, especially after 2012\cite{krizhevsky2012imagenet}. 


ConvNets are inspired on the structure of the visual cortex of the human brain, responsible for processing the visual information captured by the eye. Each layer in a ConvNet executes a simple operation, such as a weighted linear sum of the inputs, followed by a nonlinear function (e.g., the rectifier function), but when several of these layers are arranged in a deep architecture, the network can perform a highly complex and nonlinear transformation of the input data \cite{goodfellow2016deep}.

Four main concepts are explored in ConvNets: local connections, weight sharing, pooling and the use of stacked layers \cite{lecun2015deep}. 
Basically, when employed for image classification, the network process the input image pixels through different layers, performing convolution operations using shared convolution weights and extracting features from local connections. Then, these features are transformed by a nonlinear function and grouped. The output of these operations is used as input for the next layer until we reach the last layer, which essentially generates the probabilities that the input image belongs to each of the existing classes. Nowadays, due to their excellent performances, ConvNets are being applied in many real-life applications, such as object recognition and image labeling \cite{lecun2015deep}.

From the theory of Fourier transform, it is well-known that a convolution in the space domain becomes a multiplication in the frequency domain \cite{gonzalez2003digital}. So, it is perfectly possible to process the available data by working in the frequency domain, instead of performing the spatial convolution. 

This fact raises an intriguing question: what if a neural network was entirely designed to operate in the frequency domain? It is important to bear in mind that we are not referring to a standard ANN or ConvNet whose input is the frequency-domain representation of the data. Rather, we are thinking about a neural network whose operations have been, in fact, adapted to the frequency domain, in light of the theory of Fourier transform.

In this context, other important questions naturally arise, such as: what would be the model of a frequency-domain layer? How can we efficiently train such layers? Is it possible to extract both global and local frequency-domain information from the input data (images)? And, finally, what are the advantages of explicitly exploring the frequency domain, instead of using the well-established ConvNets?

The motivation for such a model comes from the idea that, depending on the problem, useful properties of the input data can be more easily perceived and extracted in the frequency domain, rather than on its raw pixels. For instance, when classifying different plantations on images captured by a drone, probably the frequency representation will bring more useful information due to the periodic nature of the data. In these images, high frequency components represent fine details, such as plant borders and corners, whereas low frequency components encode large scale information, such as shape and homogeneous areas of the plants and soil. Additionally, intermediate concentrations of frequencies in certain 2D spectrum regions will represent the periodic spacing between plants in such crops.

The present work addresses all the aforementioned questions and presents a new neural network model that explores frequency-domain layers and can be applied to different computer vision problems. The proposed model, named FreqNet, is characterized by layers of frequency filters, which extract features from the magnitude spectrum of the image, and whose parameters are adjusted with the aid of gradient-based optimization methods using the error backpropagation algorithm.

Hence, we are extending the preliminary ideas brought in \cite{stuchi_mlsp}, now presenting a technique to train the frequency-domain filters. In the previous paper, fixed frequency-domain filters were used along with a slicing procedure, which splits the image in blocks, in order to capture global and local frequency components. Here, the frequency filters for each block are fully trained using the backpropagation algorithm, allowing the proposed network to learn the most discriminative frequency components of the data and to achieve excellent results, comparable to those obtained by renowned ConvNets. Moreover, the number of adjustable weights of the network was significantly reduced by adopting a radial-based filtering approach, in which frequencies located at the same radial distance from the spectrum center share the same weight. 


It is important to remark that Fourier analysis is widely applied together with ANNs for decades and in different applications, such as brain electroencephalogram processing~\cite{srinivasan2005artificial}, cardiovascular analysis\cite{malliani1991cardiovascular}, speech processing\cite{wang2015speech}, seismic analysis~\cite{dowla1990seismic} and face spoofing detection\cite{stuchi_mlsp}. Besides, it is also explored with deep learning, as, for example, in the speech processing field~\cite{graves2013speech, silva2017exploring}, where an audio signal is converted into a 2D spectrogram image. 

However, our proposal follows a different path since it does not use handcrafted features extracted from the frequency domain, but it allows the network to learn frequency filters that boost the most discriminative components for that specific classification problem.

This paper is organized as follows: Section \ref{sec:method} describes the proposed frequency-domain architecture in detail, as well as the method used to train the network in the frequency domain. Section \ref{sec:experiments} brings the computational experiments performed in different datasets, like texture and retina images, with the purpose of evaluating the proposed method. In this section, the results and comparative analysis with deep learning architectures are also presented in order to emphasize the real contribution of the proposed method. Finally, Section \ref{sec:conc} presents the conclusions and perspectives for future research and applications.

\section{Proposed Method}\label{sec:method}

\subsection{Basic Architecture of FreqNet}

\begin{figure*}[t]
 \centering
 \includegraphics[width=1\linewidth]{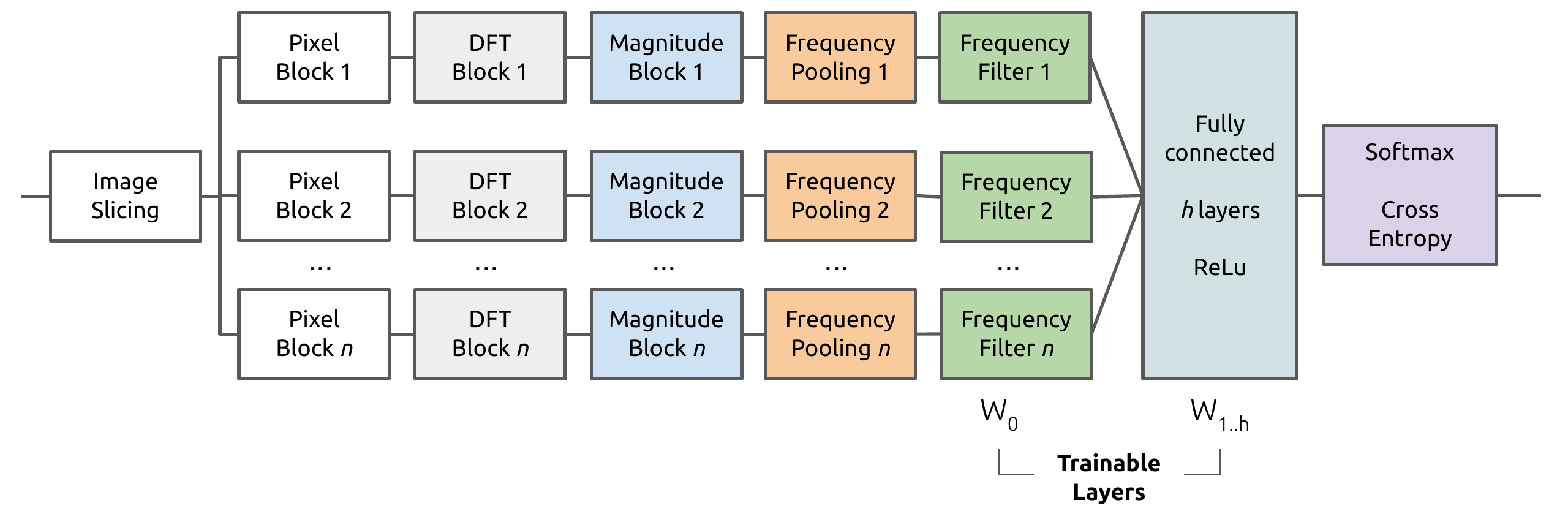}
 \caption{Basic frequency layer architecture for image classification.}
 \label{fig:architecture}
\end{figure*}

The main goal of the proposed method is to explore the frequency domain in order to improve image classification. Figure \ref{fig:architecture} presents the basic architecture behind the idea exposed in this paper.

The following parameters must be adjusted in the proposed architecture:
\begin{itemize}
    \item Image size
    \item Number of slices
    \item Frequency pooling size
    \item Frequency pooling type
    \item Number of layers
    \item Number of neurons in each layer
    \item Activation function in each layer
    \item Loss function

\end{itemize}

In addition to the aforementioned aspects, it is also necessary to choose the training algorithm and its hyperparameters (learning rate, batch size, etc).

Each step and parameter for training this architecture will be presented in the sequence.

\subsection{Image slicing}

The first step consists in slicing the input image in several blocks. The purpose of this step is to allow the network to extract both global and local information. When the frequency spectrum is calculated for the entire image, global features of this image can be extracted; on the other hand, when the image is sliced in small blocks, these will provide local features. Examples of global features are the background and main object components, while examples of local features are small textures, corners and borders.

\begin{figure}[!htbp]
 \centering
 \includegraphics[width=1\linewidth]{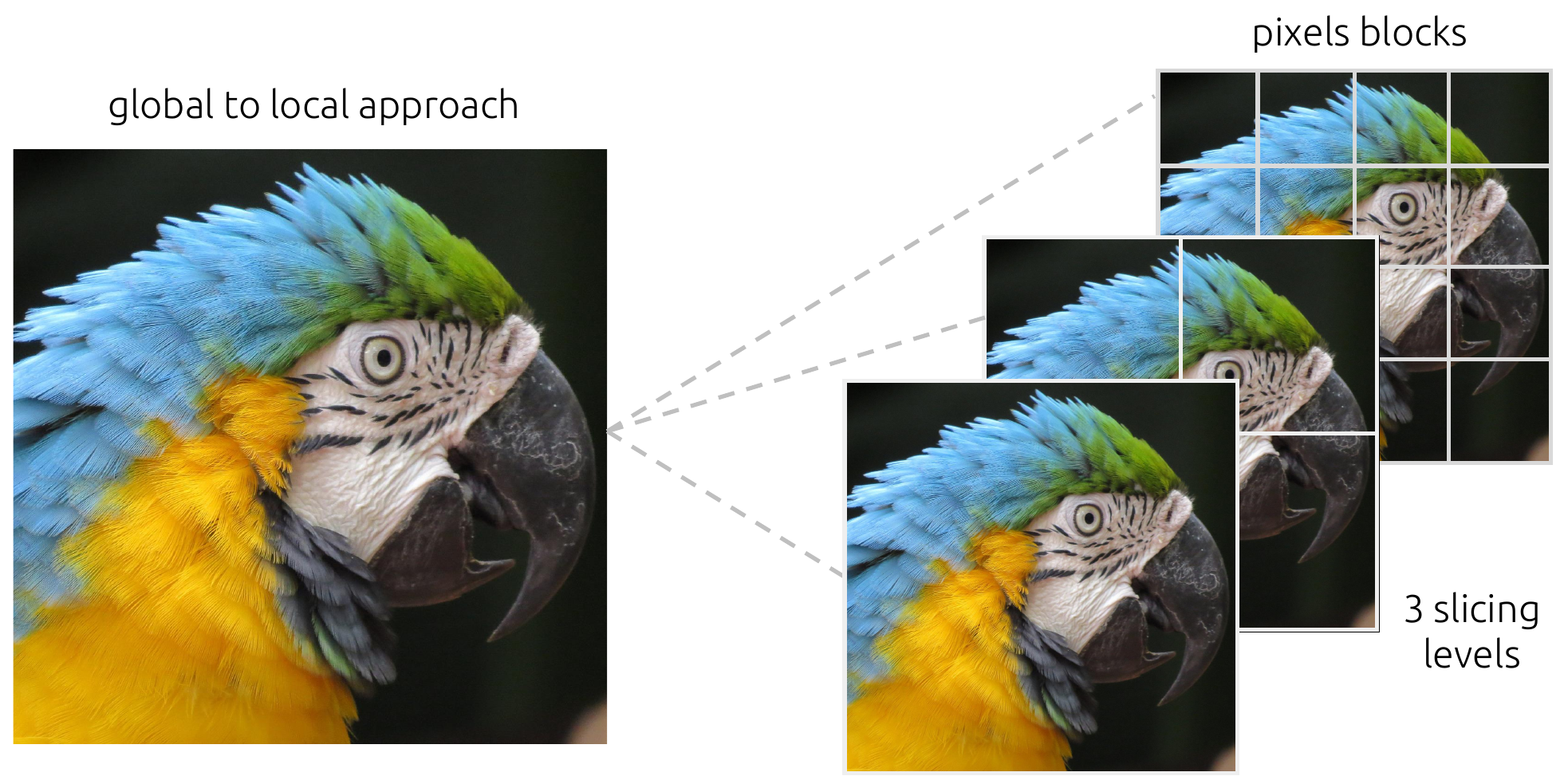}
 \caption{Basic slicing process.}
 \label{fig:slice1}
\end{figure}

Figure \ref{fig:slice1} shows the basic slicing process. For the sake of clarity, three slicing levels are shown in this figure. Basically, in the first level the whole original image is considered.
In the second level, the original image is divided in 4 blocks, each one with half of the original size. In the third level, each of the four blocks from the previous slicing is divided in four new blocks again, thus yielding a total of 16 blocks. This process can be repeated until the chosen maximum number of slicing levels is attained. 

\begin{figure}[!htbp]
 \centering
 \includegraphics[width=1\linewidth]{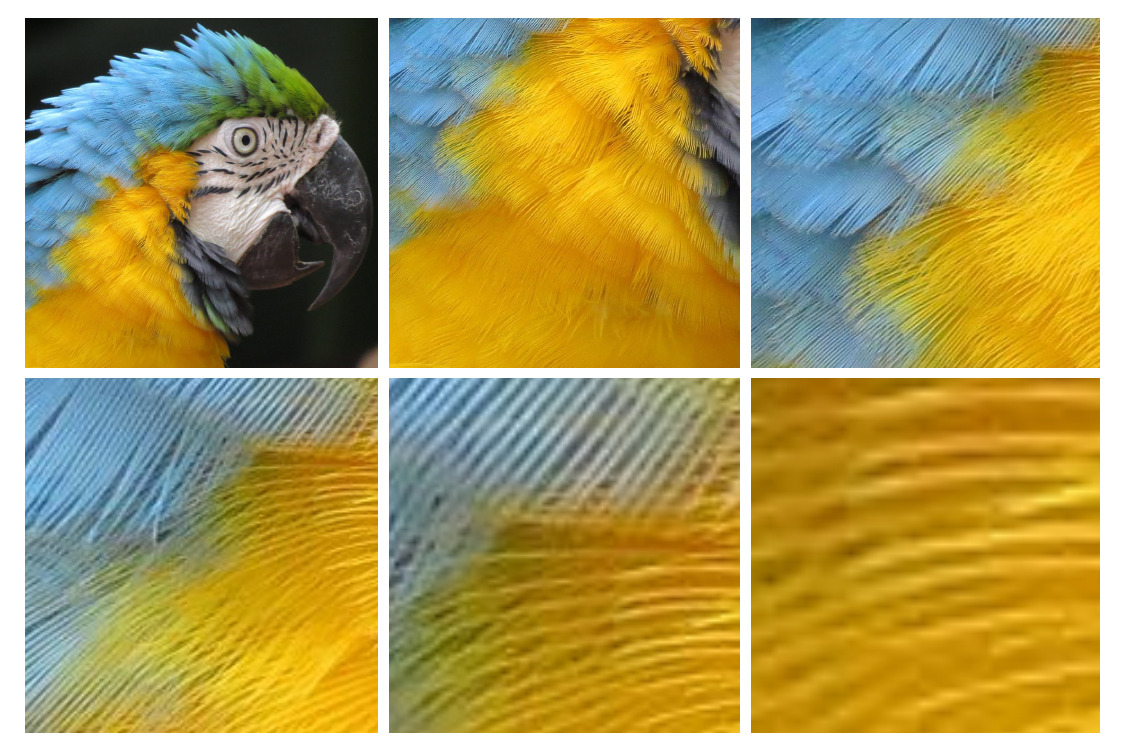}
 \caption{Blocks from six slicing levels from macaw image showing different frequencies in the feathers.}
 \label{fig:slice2}
\end{figure}

An example of the slicing blocks is shown in Figure \ref{fig:slice2}, where it is possible to see more details of the bird feathers as the level increases, similarly to a zooming in effect in the image. Bigger blocks bring global information, as the low frequencies of the background and the main macaw components. On ther other hand, small blocks represent the local information, like the specific textures in the feather of this bird species. It is worth mentioning that even looking at the small blocks, it is possible to recognize that this could be part of a macaw. The proposed algorithm explores this possibility since it will discover the pertinent frequencies that should be emphasized during the training process.

\subsection{2D Discrete Fourier Transform}

After the slicing, each block will be transformed to the frequency domain. 







In this step, the 2D Discrete Fourier Transform (DFT)~\cite{gonzalez2003digital} is employed. For each image block $i$, denoted as $f_i(x,y)$, with size $A_i \times B_i$, the DFT is computed according to the following expression: 

\begin{equation}
F_i(u, v)=\sum_{x=0}^{A_i-1}\sum_{y=0}^{B_i-1} f_i(x, y)e^{-j2\pi(ux/A_i+vy/B_i)}
\end{equation}

\begin{figure}[!htbp]
 \centering
 \includegraphics[width=1\linewidth]{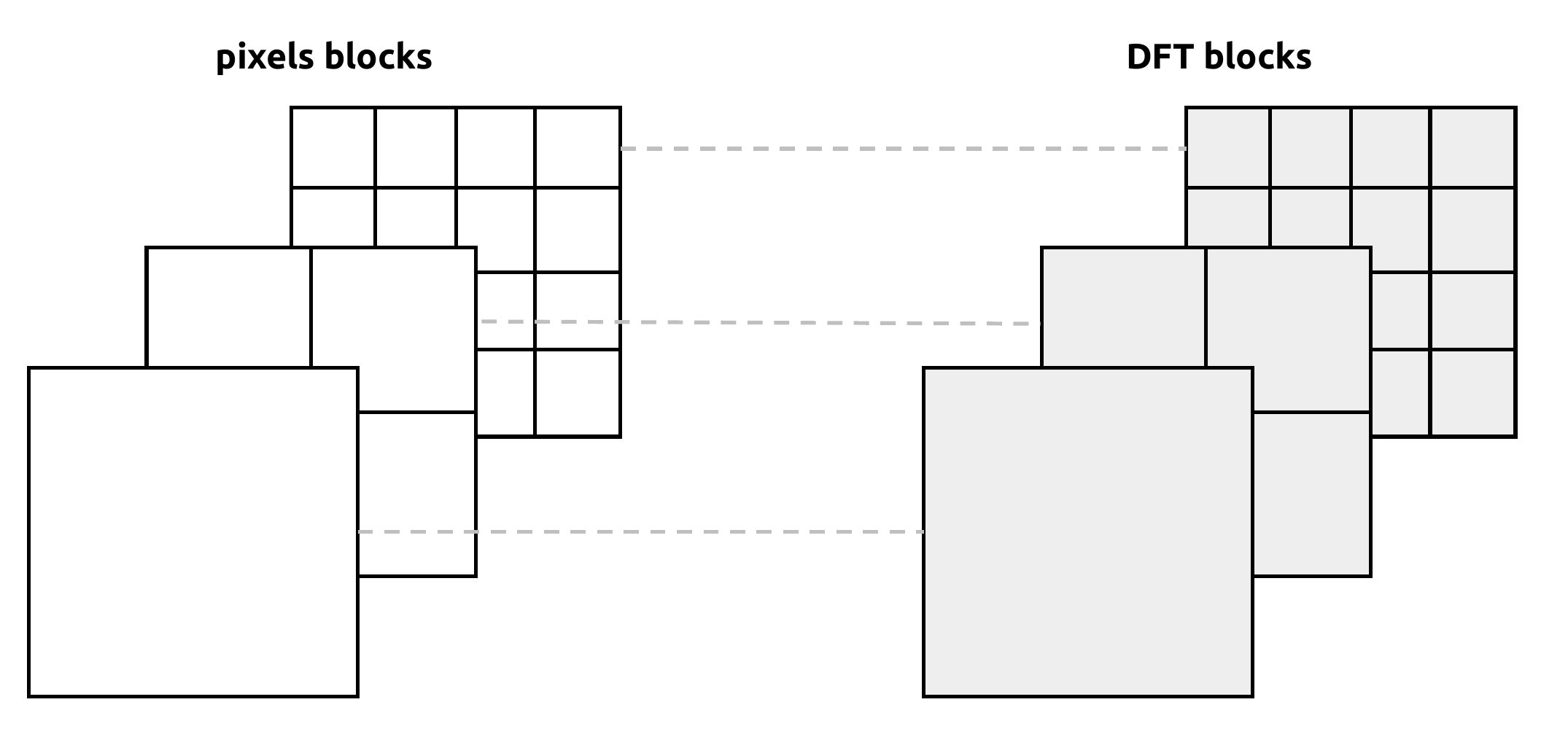}
 \caption{Discrete Fourier Transform for each image block.}
 \label{fig:dft}
\end{figure}

Figure \ref{fig:dft} depicts this step. In this case, considering three slicing levels, 21 DFTs need to be calculated, being one for the entire image, four for the second level and sixteen for the last level. 

\begin{figure}[!htbp]
 \centering
 \includegraphics[width=1\linewidth]{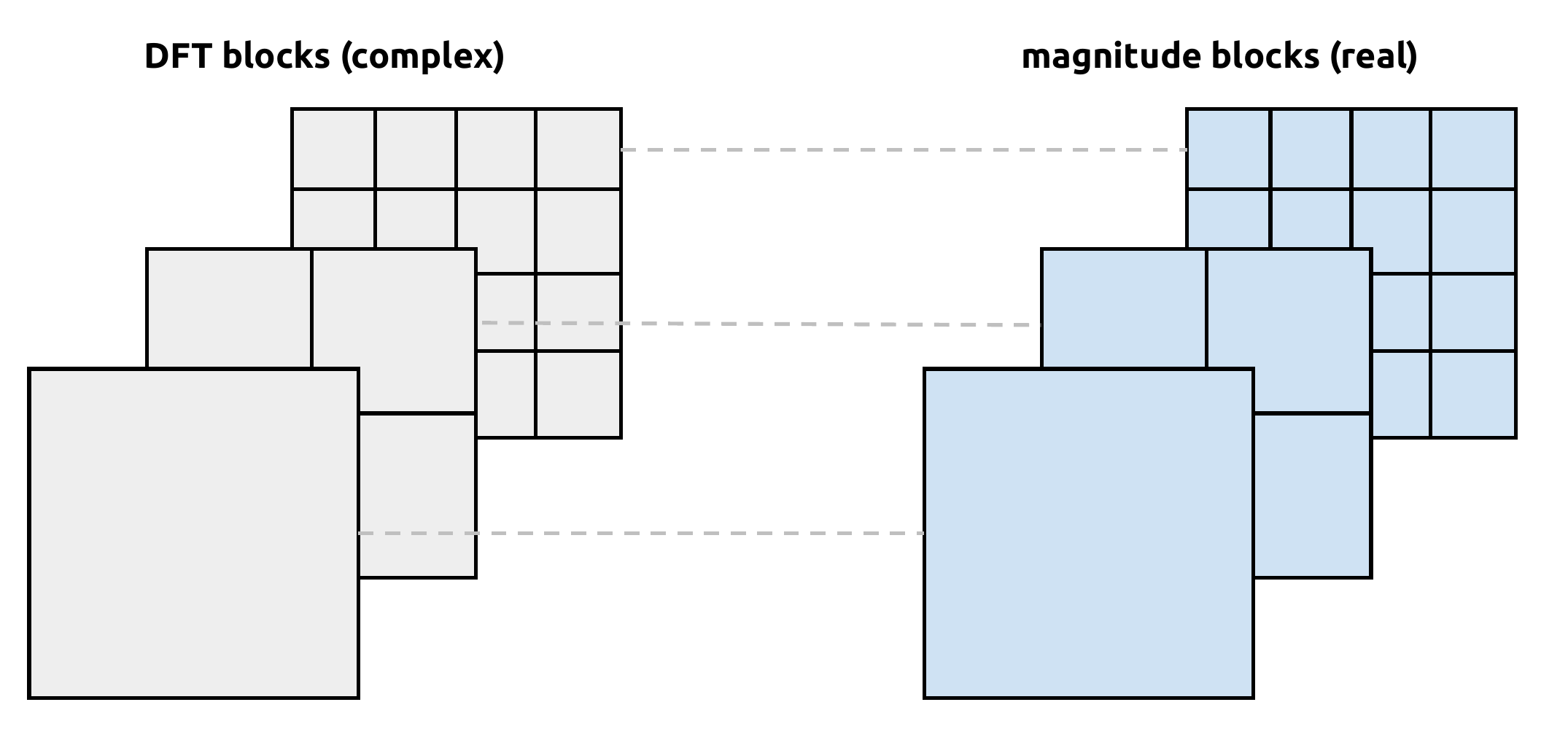}
 \caption{DFT magnitude calculated for each image block.}
 \label{fig:magnitude}
\end{figure}

Since the DFT yields a matrix of complex numbers and most ANN/DL operations deal with real numbers, the magnitude of the spectrum of each block is obtained, according to Equation (2), and effectively explored by the network. This step is represented by the Figure \ref{fig:magnitude}.

\begin{equation}
M_i(u, v) = \sqrt {\left( {Real(F_i(u, v) } \right)^2 + \left( {Imag(F_i(u, v)} \right)^2 }
\end{equation}

\subsection{Frequency pooling}
\label{sec:freq_pool}

In the DFT domain, the frequencies radially vary from low, in the center of the spectrum, to the high, in the borders. In this context, frequencies are commonly filtered in a radial way. For example, a band-pass filter can be designed in the shape of a ring, allowing filtering specific frequencies from the image.

\begin{figure}[!htbp]
 \centering
 \includegraphics[width=1\linewidth]{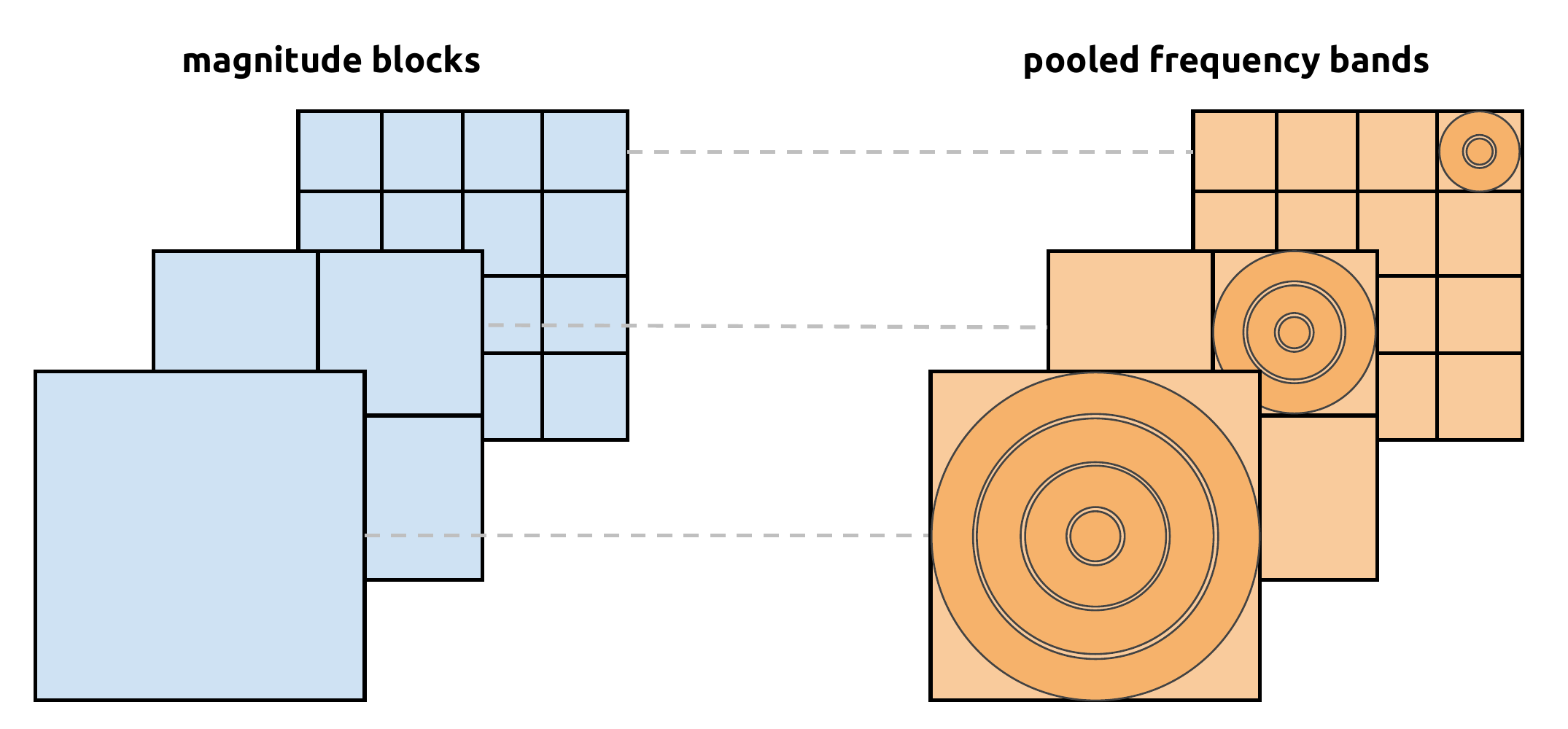}
 \caption{Pooling in frequency to reduce data dimensionality.}
 \label{fig:freq_pooling}
\end{figure}

In this sense, frequency components associated with close radii can be grouped, if desired. In other words, we may choose to process larger frequency bands instead of single cells sharing the same radius. As shall be described in Section \ref{sec:freq_filt}, the width of the frequency bands directly influences the number of filter weights that need to be tuned by the network. This pooling step is shown in Figure \ref{fig:freq_pooling}.

\begin{figure}[!htbp]
 \centering
 \includegraphics[width=0.75\linewidth]{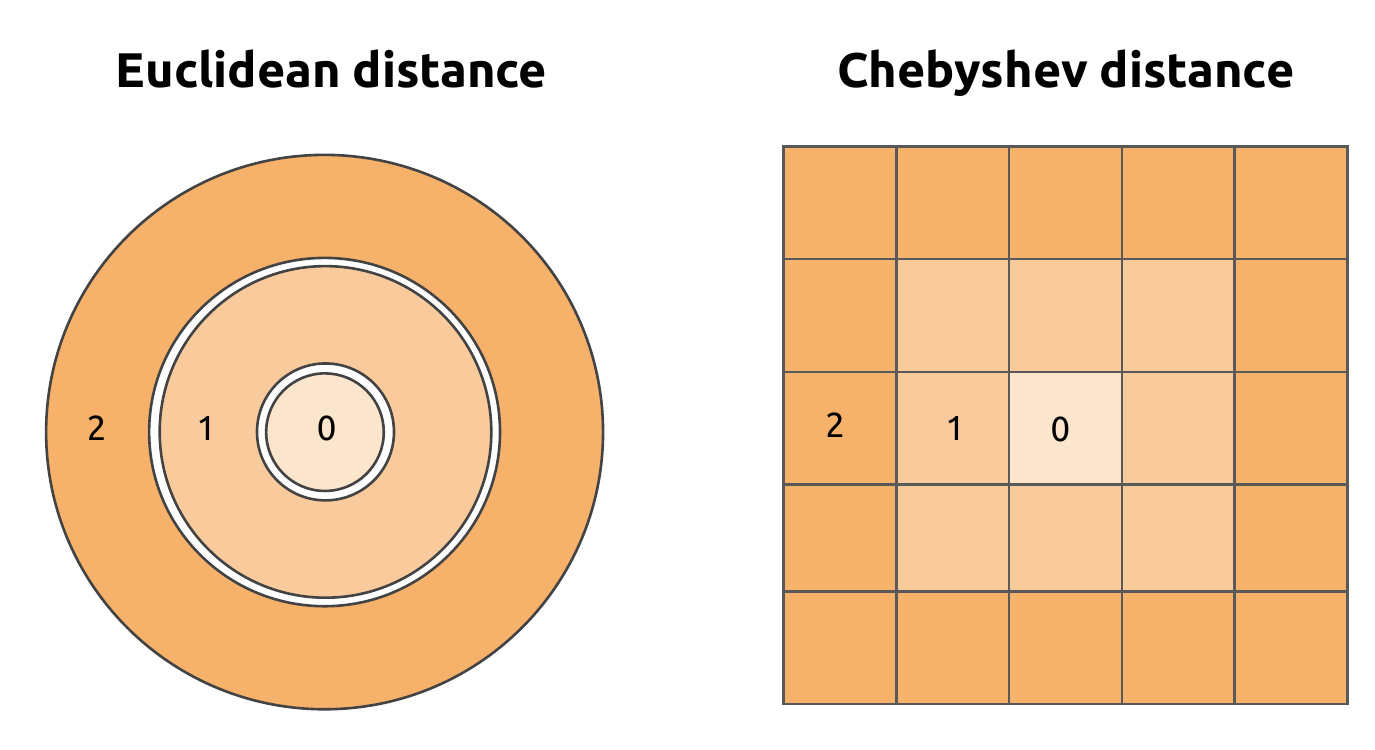}
 \caption{Euclidean or Chebyshev distances can be used to group frequencies in the spectrum.}
 \label{fig:distance}
\end{figure}

Analogously to the case of radial bands, it is also possible to adopt squared frequency groups. 
In this case, the only difference is that the distance between a particular cell and another in the center of the image is calculated using Chebyshev distance, instead of Euclidean\cite{webb2003statistical}. Figure \ref{fig:distance} shows the difference between them.


The Chebyshev distance grouping can be convenient in this pooling step since it simplifies how the highest frequencies are aggregated, since the maximum circle that can be defined within the magnitude block loses high frequencies in the spectrum borders when using Euclidean distance. Option for radial or squared frequency grouping establishes the pooling type. 

Therefore, this operation is useful to reduce the dimensionality for the next steps, simplifying and reducing the time spent in the training process.

\subsection{Frequency Filtering}
\label{sec:freq_filt}

\begin{figure}[!htbp]
 \centering
 \includegraphics[width=1\linewidth]{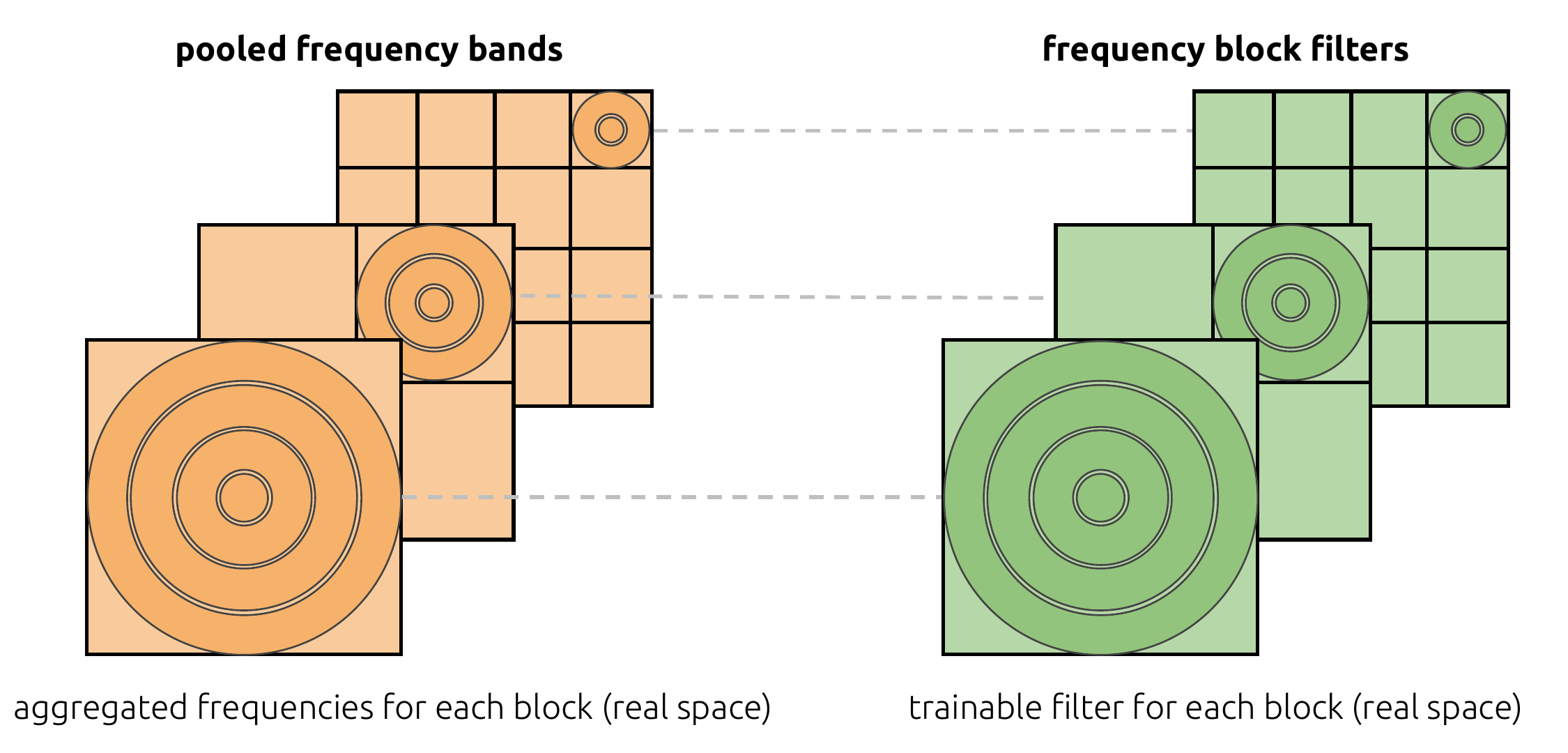}
 \caption{Frequency filtering for each block.}
 \label{fig:filtering1}
\end{figure}

After these three steps, the first trainable layer is presented. In this layer, the pooled frequency magnitude will be filtered by weights that will be learned during the training process. In this case, for filtering in the frequency domain, the magnitude will be element-wise multiplied by a new matrix $W_i \in R^{2}$ , which represents the filter.  Figure \ref{fig:filtering1} represents this filtering step.

\begin{figure}[!htbp]
 \centering
 \includegraphics[width=1\linewidth]{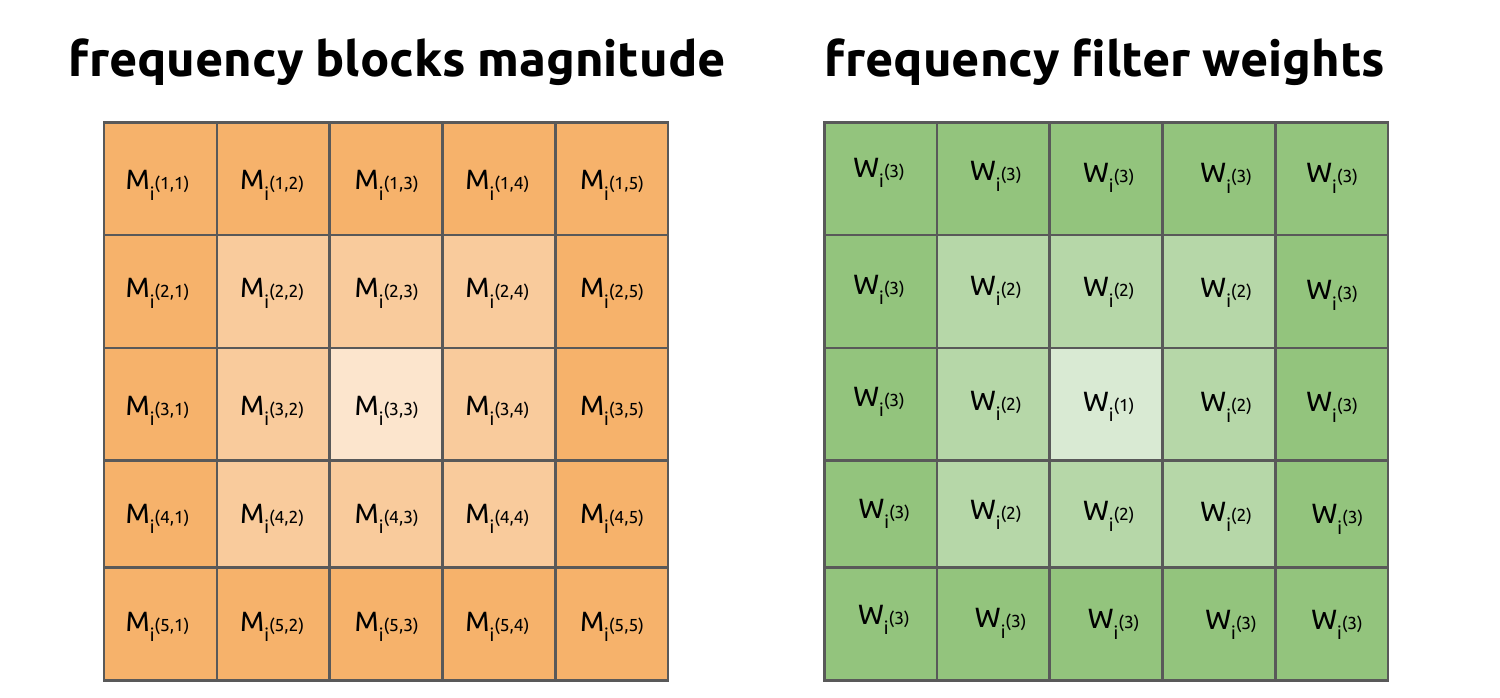}
 \caption{Frequency filtering simplification.}
 \label{fig:filtering3}
\end{figure}

In order to simplify the filtering processing, it is possible to define weighted rings or squares, according to the selected pooling type, which will filter each block in the frequency domain. 
In this case, even though the filter matrix $W_i$ has the same size of the image block, only a few weights need to be adjusted, each representing one radius, as shown in Figure \ref{fig:filtering3}. Additionally, as remarked in Section \ref{sec:freq_pool}, if pooling is adopted, the number of weights can be further reduced, as each filter weight is responsible for processing a wider frequency band. 


In Figure \ref{fig:filtering3} the magnitude filtering will be performed by the element-wise multiplication of the magnitude ($5\times5$ orange matrix on the left) by the frequency filter ($5\times5$ green matrix on the right). Notice that only three filter weights will be learned on the training step, being: $W_i(1)$, $W_i(2)$ and $W_i(3)$.
This process is applied for each block $i$, multiplying its respective magnitude $M_i$ by the frequency weights $W_i$, as follows:

\begin{equation}
\mu_i = M_i \odot W_i
\label{eq:magnitude_filter}
\end{equation}

After this multiplication, a coefficient $C_i(r)$ is extracted from each ring with radius $r$ and for each block $i$. Let $R$ be the maximum distance from the center, which is half of the block size. Then, the coefficient $C_i(r)$ amounts to the sum of the elements inside the group with radius $r$, as shown in the following expression:


\begin{equation}
C_i(r)= \sum_{r=1}^{R} \mu_i(r)
\label{eq:coefficient_calculation}
\end{equation}

Using this idea, a substantial reduction in the amount of coefficients to train is achieved. For example, with an image of $1024 \times 1024$ pixels, there would be $1.048.576$ weights to be adapted for the first slicing level. However, using the proposed simplification, only $512$ coefficients will be trained, which makes the training process much faster and feasible. 


\subsection{Fully connected training}

\begin{figure}[!htbp]
 \centering
 \includegraphics[width=1\linewidth]{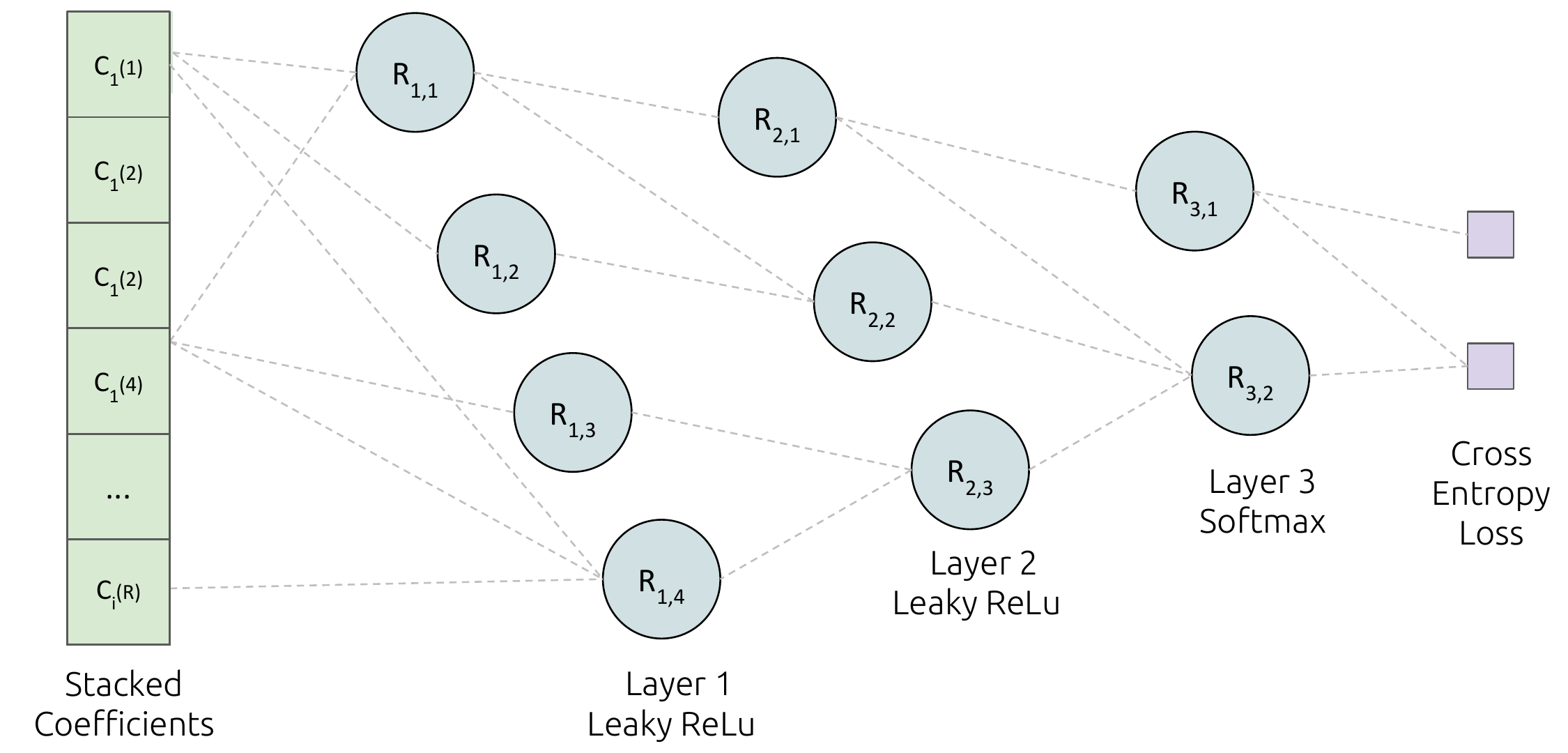}
 \caption{Coefficients stacked as input for the fully connected network.}
 \label{fig:fully_connected}
\end{figure}

After the frequency filtering step, all the coefficients $C_i(r)$ are stacked and used as inputs for the fully connected layer. Figure \ref{fig:fully_connected} shows this step.

Different number of layers, neurons, activation and loss functions can be applied as in a regular fully connected architecture. In the approach presented here, all the hidden layers explore \textit{leaky ReLu} activation, while the output layer is a \textit{softmax} layer, thus yielding the probability for each class. Then, the \textit{cross-entropy} is adopted as the loss function of the model to allow image classification. 


It is important to highlight that training the FreqNet model involves the adaptation of the weights of the fully connected layers (and \textit{softmax} layer), as well as of the filter weights in the frequency filtering stage. Interestingly, the frequency filtering operations, defined in Equations (\ref{eq:magnitude_filter}) and (\ref{eq:coefficient_calculation}) easily allow the computation of the gradient with respect to the weights. Hence, the training process of FreqNet backpropagates the error until the frequency layer and updates the weights of the frequency filters, as well as the parameters of the fully connected network. 

Since we are processing the magnitude of the spectrum, what is learned by the network is the magnitude spectrum of the filters. Therefore, the weights $W_i (j)$ cannot be negative. A simple way to guarantee this condition consists in truncating to zero the potential negative weights that might arise after the weight update for each mini-batch. Interestingly, this is similar to applying the rectifier function over the filter weights.

By exploring the proposed architecture, with all the ideas and stages already described, interesting results have been obtained. The methodology and experimental results shall be presented in the next section. 

\section{Experiments, Results and Discussion}
\label{sec:experiments}

In order to evaluate the proposed method, different datasets were used. The first one is based on textures from Kylberg dataset \cite{Kylberg2011c}. After the analysis on texture classification, we considered retina datasets for two distinct tasks: (1) detection of normal and cataract patients\cite{kaggle_cataract}; and (2) classification of the retina image quality\cite{eyeq}. The computational experiments will be presented in the sequence.

The machine used for training all the models is an Intel i7-8565U CPU with four $1.80$GHz cores, Nvidia GeForce MX130 GPU, $8$GB of RAM memory, $256$GB of SSD memory and $2$TB of hard disk.

\subsection{Texture Experiments}
\subsubsection{Kylberg Dataset}

\begin{figure}[!htbp]
 \centering
 \includegraphics[width=1\linewidth]{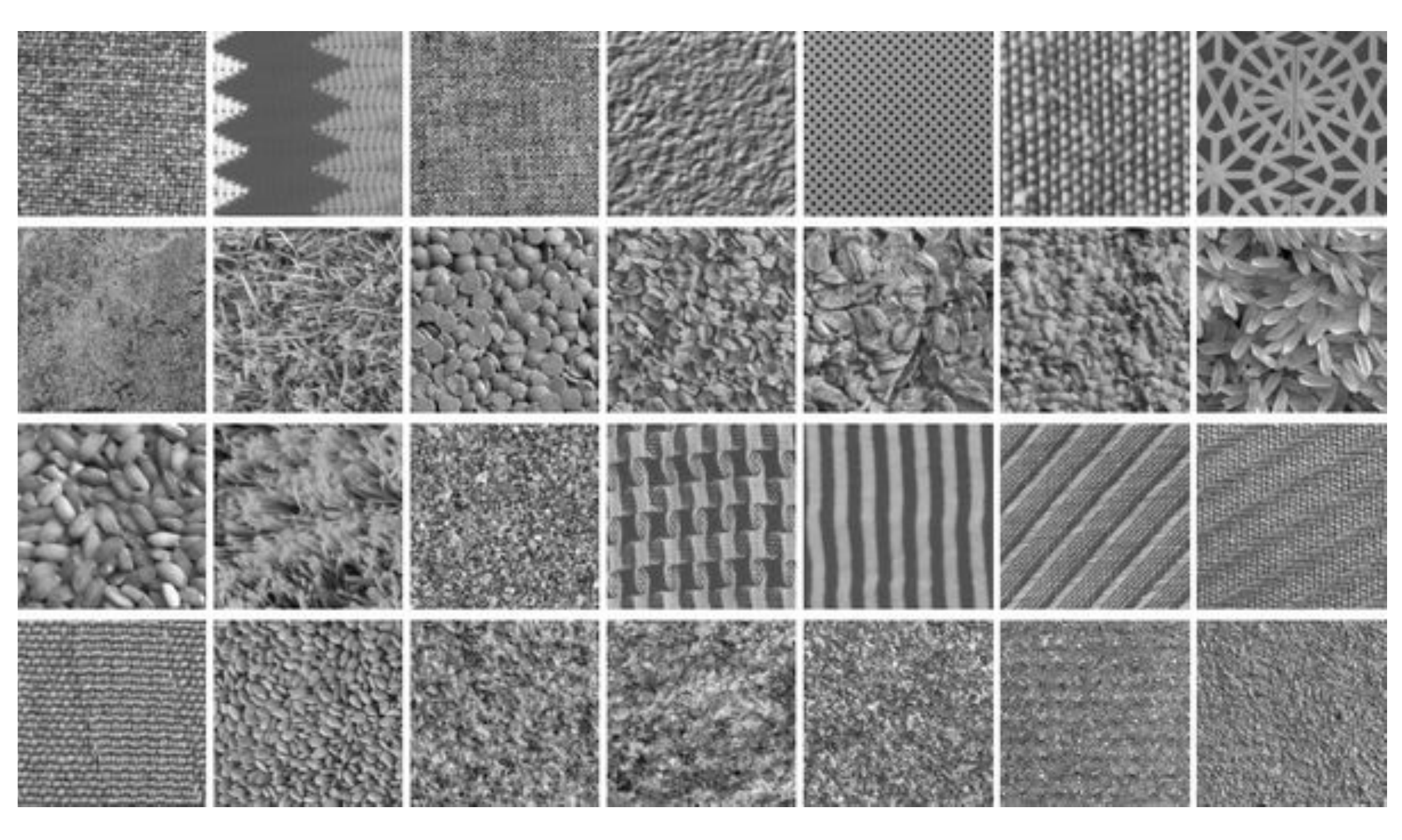}
 \caption{Examples of each class of the Kylberg dataset\cite{Kylberg2011c}.}
 \label{fig:kylberg}
\end{figure}

The Kylberg texture dataset is composed of images from 28 textures classes, such as rice, stone, lentils, rug and wall. Figure \ref{fig:kylberg} shows one example of each class.

\begin{figure}[!htbp]
 \centering
 \includegraphics[width=0.75\linewidth]{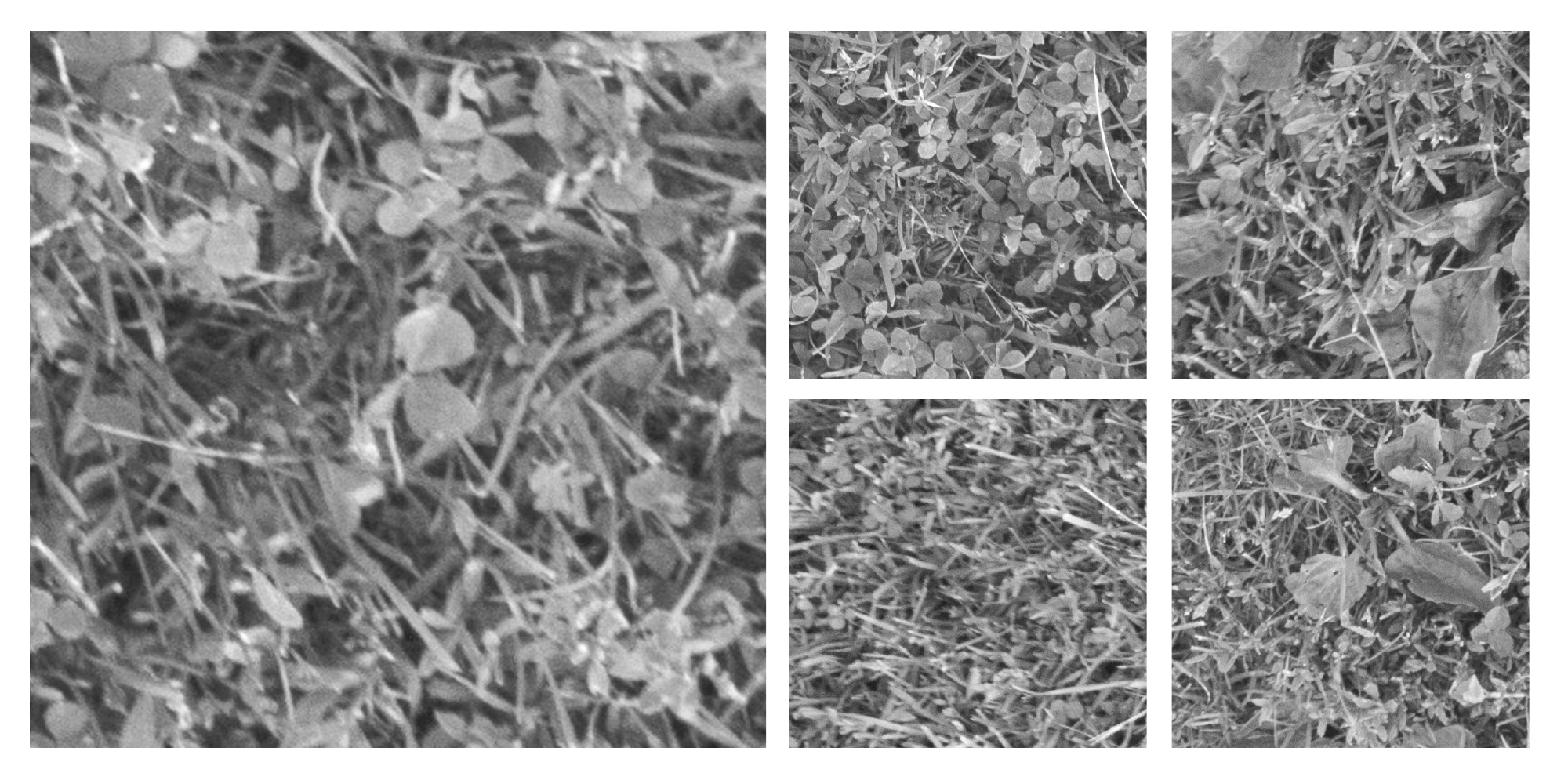}
 \caption{Example of \textit{grass} class from Kylberg texture dataset.}
 \label{fig:grass}
\end{figure}

Each image has $576\times576$ pixels lossless compressed in 8-bit PNG format. Each class has 160 unique images, resulting in 4480 images. Figure \ref{fig:grass} shows some examples of the \textit{grass} class, where it is possible to see that the textures vary based on the place where the image was captured. In the case of the \textit{grass} class, different types of plants are present in the image.

\begin{figure}[!htbp]
 \centering
 \includegraphics[width=0.75\linewidth]{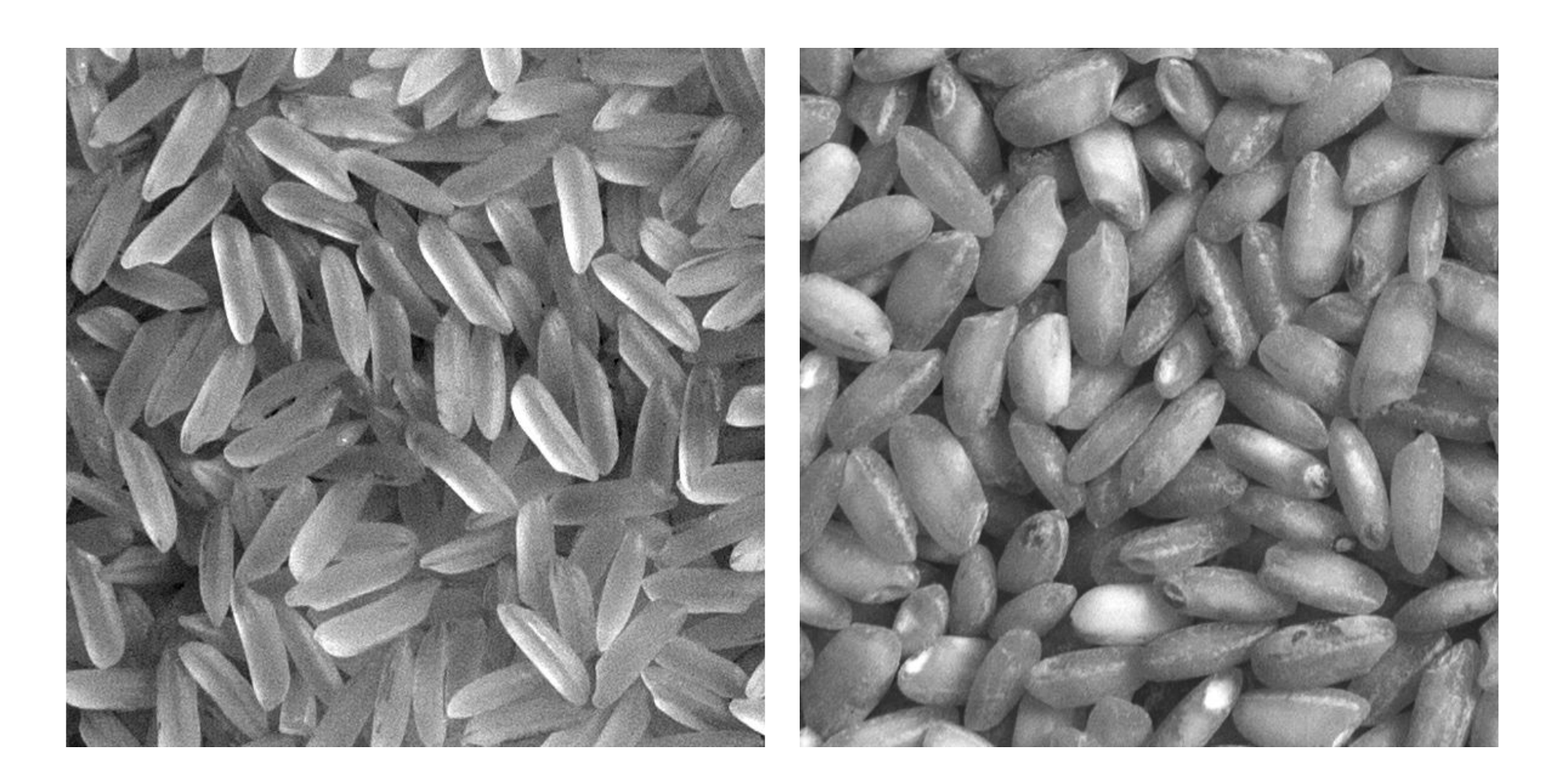}
 \caption{Example of \textit{rice 1} and \textit{rice 2} classes from Kylberg dataset.}
 \label{fig:rice}
\end{figure}

The dataset also contains classes from closer categories, such as rices from different species, named \textit{rice 1} and \textit{rice 2}. Figure \ref{fig:rice} exhibits one example of these two types of rice. 

\begin{figure}[!htbp]
 \centering
 \includegraphics[width=0.75\linewidth]{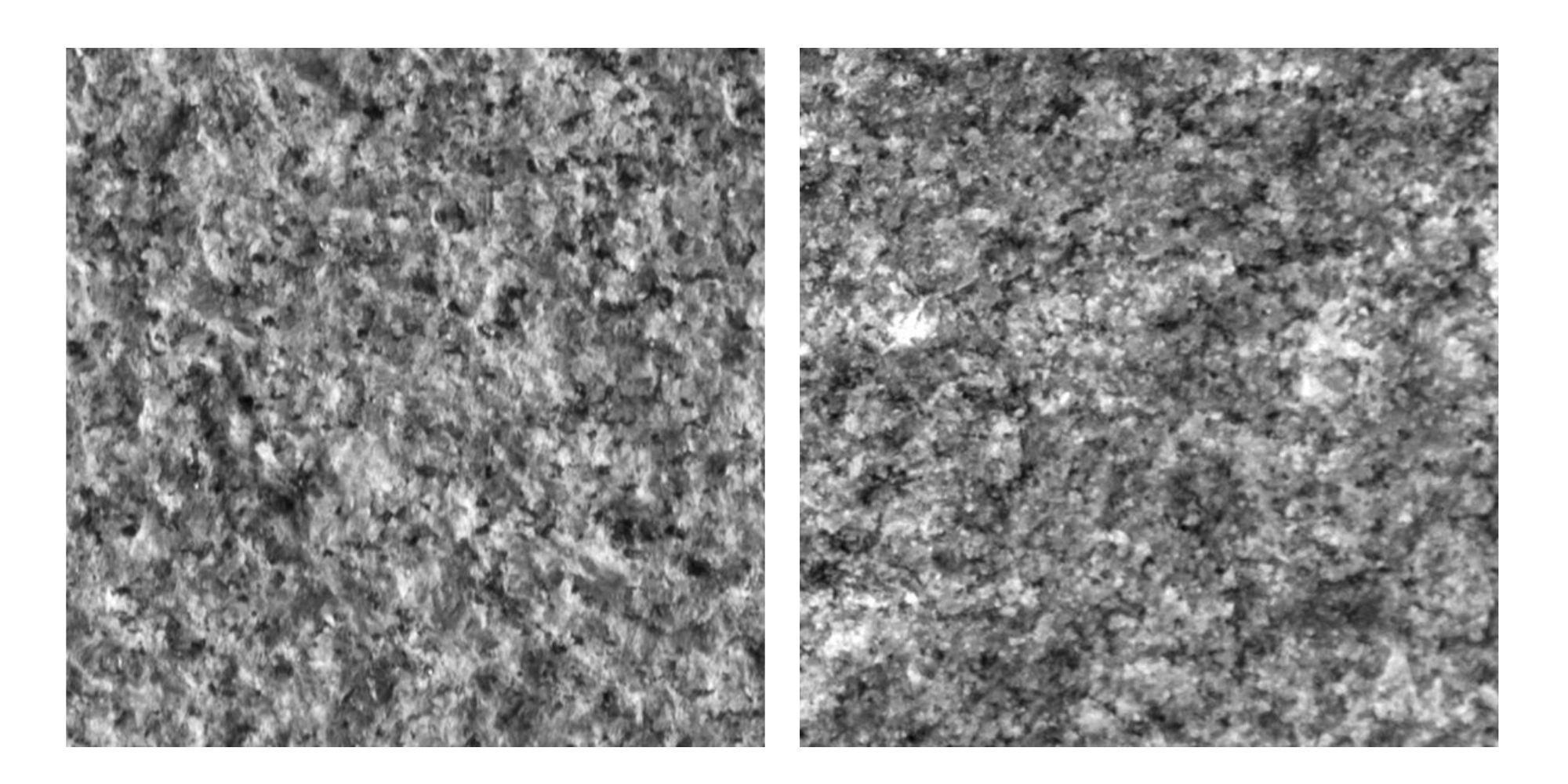}
 \caption{Example of \textit{stone 1} and \textit{stone 2} classes from Kylberg dataset.}
 \label{fig:stone}
\end{figure}

Another pair of classes that poses a significant challenge for the classification is related to \textit{stone 1} and \textit{stone 2} classes, as presented in Figure \ref{fig:stone}. As we can observe, images from these two classes can be quite similar, so that it is hard to correctly distinguish them, even for a human reviewer.  

Some experiments have been carried out using this dataset, which are described in the next subsections. 

\subsubsection{Experiment I - 2 textures classification}

\begin{figure}[!htbp]
 \centering
 \includegraphics[width=0.75\linewidth]{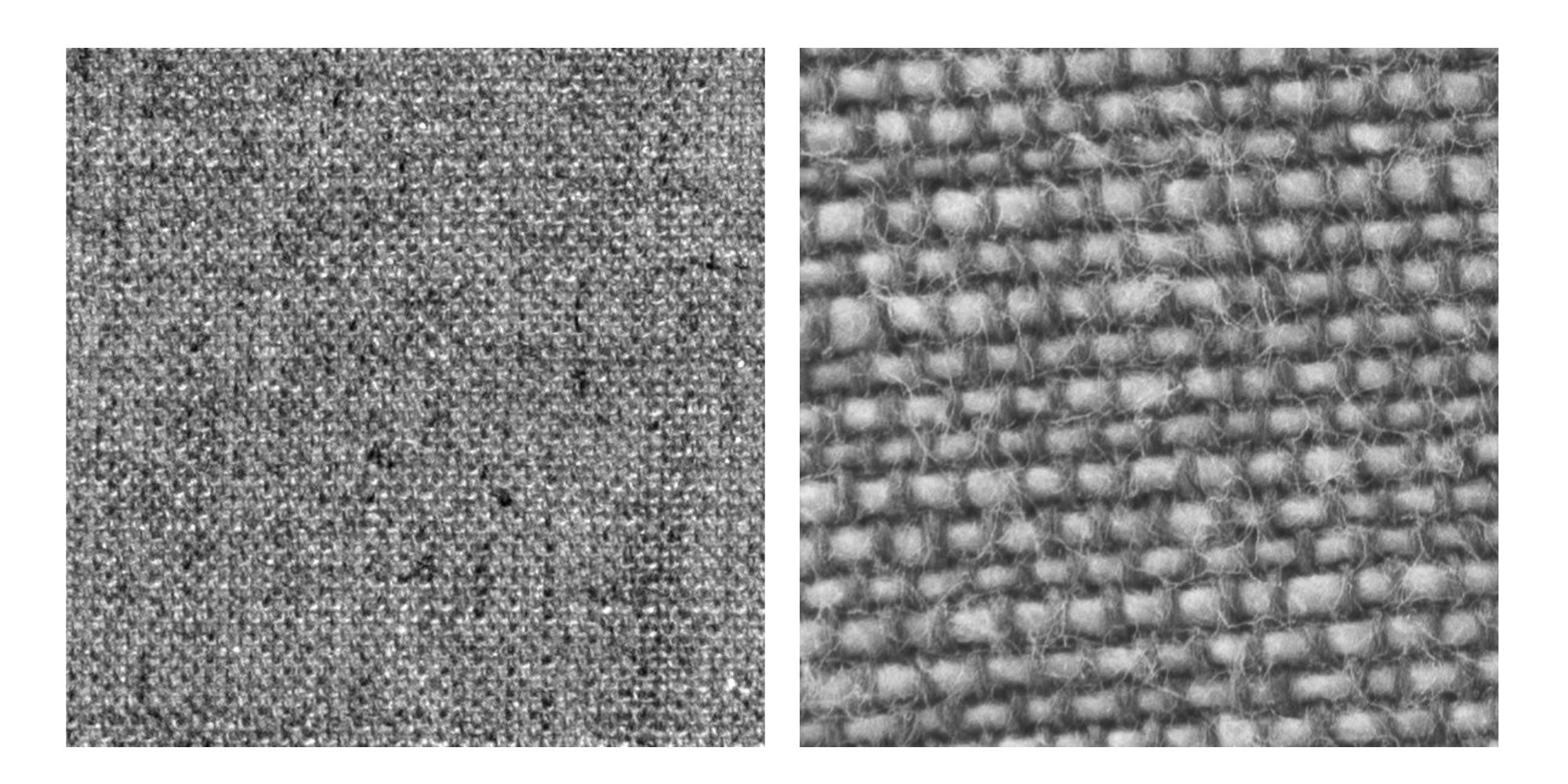}
 \caption{Example of \textit{canvas} and \textit{cushion} classes from Kylberg dataset.}
 \label{fig:canvas_cushion}
\end{figure}

The first scenario was designed to validate the proposed method and to verify the separation of two classes with distinct patterns in the frequency domain. For this purpose, the \textit{canvas} and \textit{cushion} classes were selected from Kylberg dataset. Figure \ref{fig:canvas_cushion} shows one example from these two classes.

\begin{figure}[!htbp]
 \centering
 \includegraphics[width=0.75\linewidth]{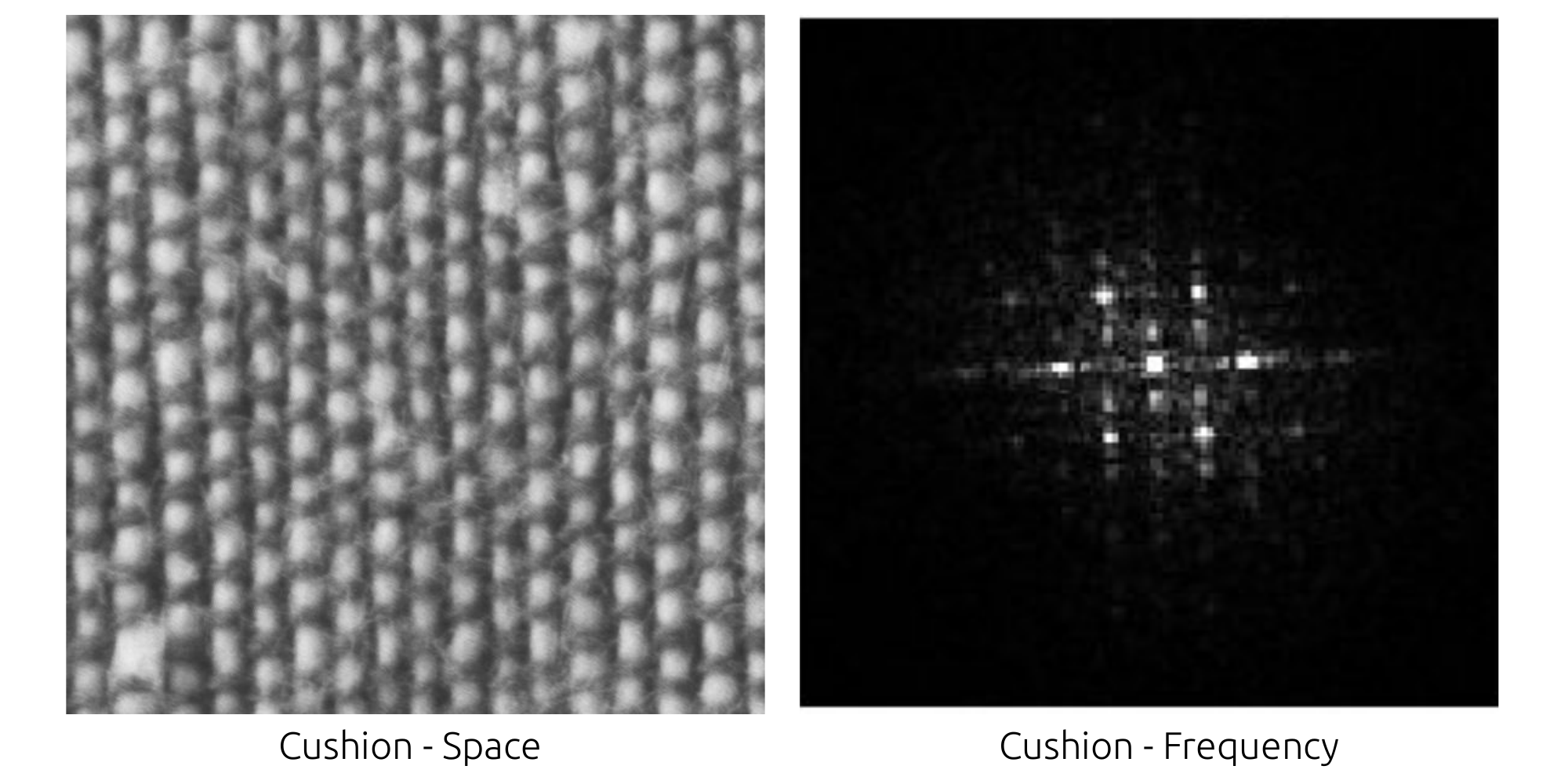}
 \caption{Example of \textit{cushion} and the corresponding DFT (magnitude).}
 \label{cushion_frequency}
\end{figure}

It is possible to notice that the classes are formed by different fabric threads, which leads to different behaviors in the frequency domain. 
In this experiment, the images were resized for $128\times128$ pixels and the 2D-DFT for the \textit{cushion} and \textit{canvas} classes were calculated. Figure \ref{cushion_frequency} shows the DFT magnitude for \textit{cushion} class, where it is possible to see that the main frequencies are closer to the center.

\begin{figure}[!htbp]
 \centering
 \includegraphics[width=0.75\linewidth]{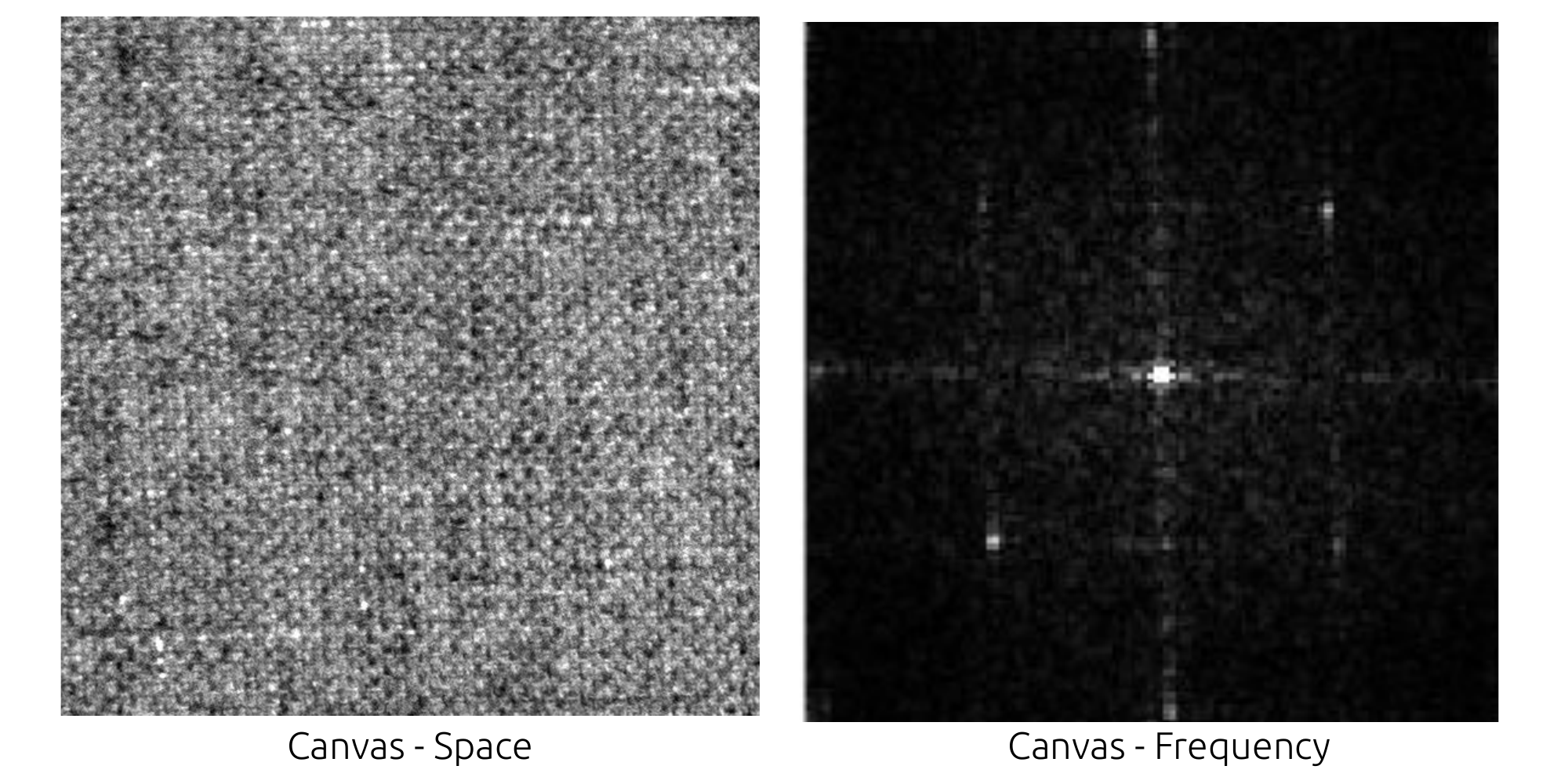}
 \caption{Example of \textit{canvas} and the corresponding DFT (magnitude).}
 \label{fig:canvas_frequency}
\end{figure}

On the other hand, for the \textit{canvas} image, it is possible to notice that the main frequency components are more distant to the spectrum center, as shown in Figure \ref{fig:canvas_frequency}.  

\begin{figure*}[!htbp]
 \centering
 \includegraphics[width=1\linewidth]{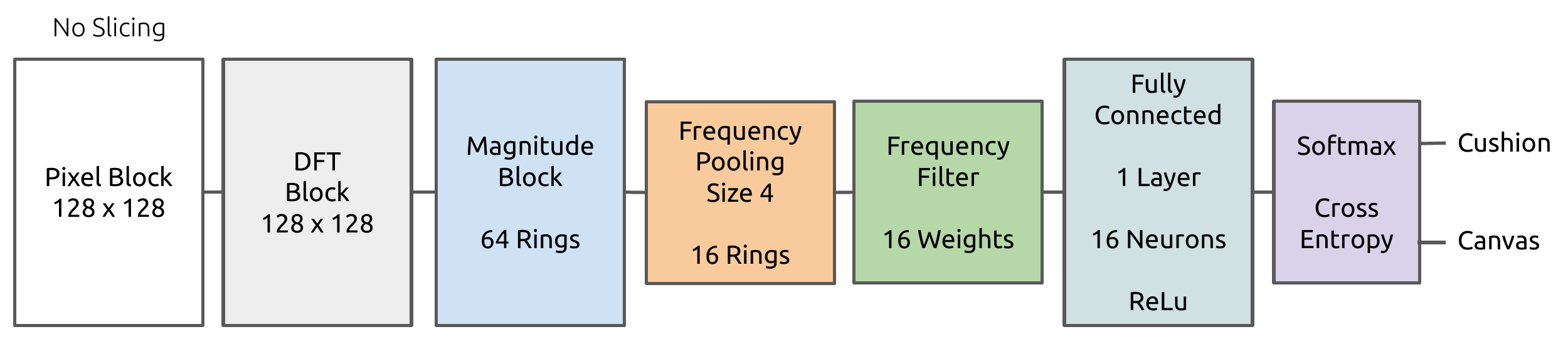}
 \caption{Architecture used to classify \textit{canvas} and \textit{cushion} classes.}
 \label{fig:freq_net_cushion_canvas}
\end{figure*}

\begin{figure}[!htbp]
 \centering
 \includegraphics[width=1\linewidth]{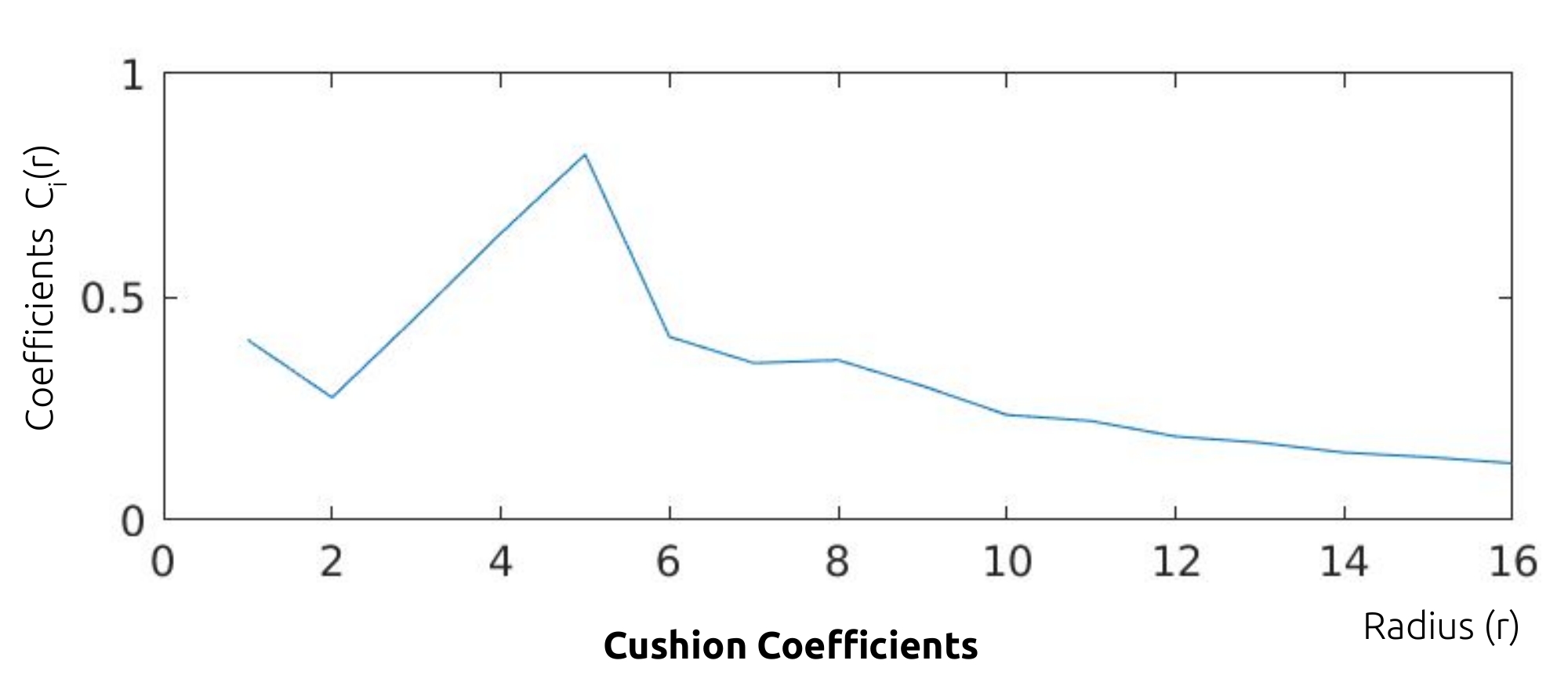}
 \caption{Example of \textit{cushion} coefficients extracted from frequency domain.}
 \label{fig:cushion_coefficients}
\end{figure}

For these $128\times128$ pixels spectrum matrices, $64$ rings can be formed radially from the center to the borders. By adopting a frequency pooling with size $4$, $16$ rings were created. 
Then, by summing the magnitudes of all frequency pixels within each ring for the \textit{cushion} class, we obtained a set of coefficients similar to the one presented in Figure \ref{fig:cushion_coefficients}. As we can observe, there is a peak around radius $5$ in the pooled frequency magnitude. 

\begin{figure}[!htbp]
 \centering
 \includegraphics[width=1\linewidth]{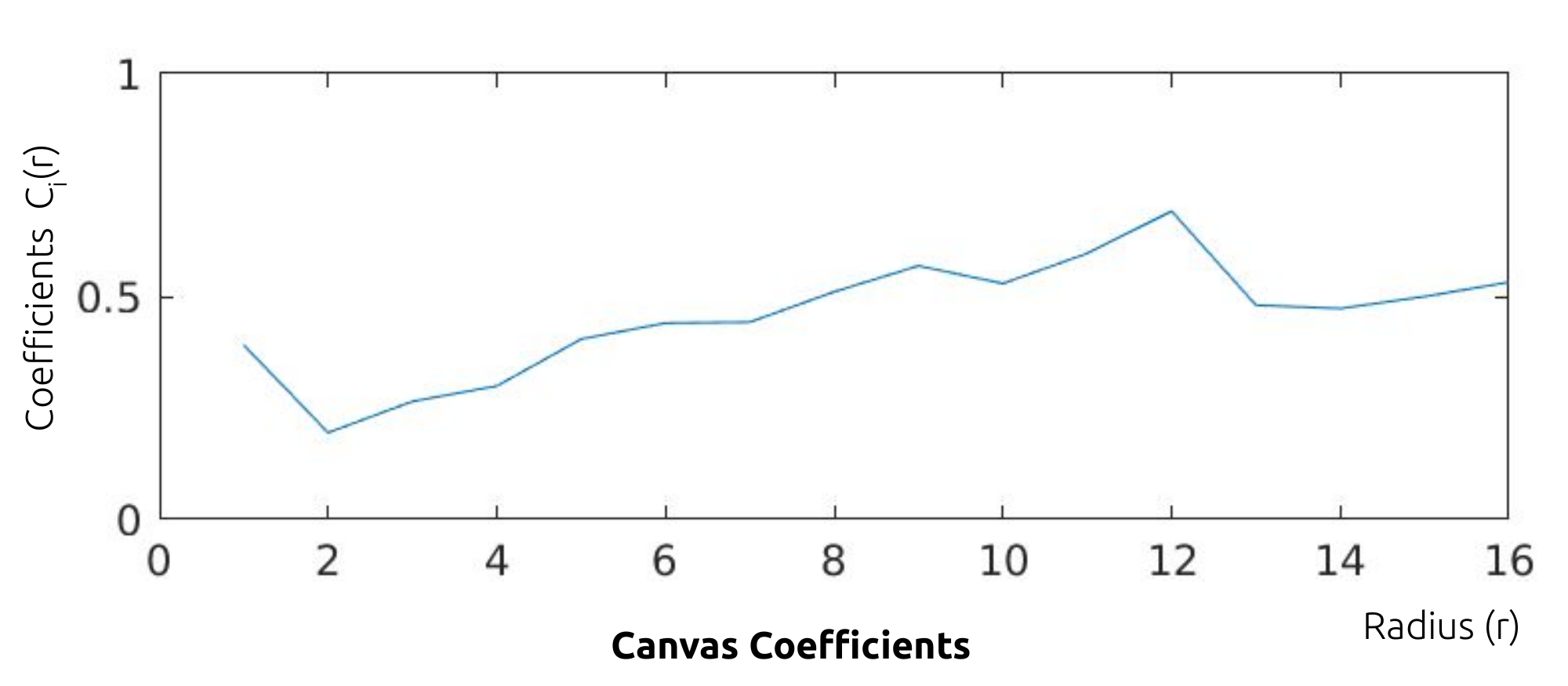}
 \caption{Example of \textit{canvas} coefficients extracted from frequency domain.}
 \label{fig:canvas_coefficients}
\end{figure}

For the \textit{canvas} class the same operation was performed. In this case, the pooled frequency magnitude contains a peak around the radius $12$, as can be seen in Figure \ref{fig:canvas_coefficients}. 

Therefore, the set of frequency-domain coefficients extracted for \textit{cushion} and \textit{canvas} classes exhibit a distinct behavior as a function of the radius, which means that different frequency bands are more pronounced in \textit{cushion} and \textit{canvas} images. So, it is plausible to expect that the frequency filters will emphasize those two spectrum bands, where the classes clearly have distinct behaviors, in order to separate them. This aspect was the main motivation for this first experiment. The architecture used for the classification of these two classes is presented in Figure \ref{fig:freq_net_cushion_canvas}.

Since the problem is relatively simple to be solved by a classifier, no image slicing was applied, since the global frequencies are sufficient to distinguish these two classes. So, the $320$ images from these two classes were split in training and validation sets, using $75\%$ and $25\%$ respectively. 
After that, the 2D-DFT was calculated for each image. In the sequence, the spectrum magnitude was calculated and the frequencies were pooled in rings with size $4$, resulting in $16$ rings. Each frequency ring is then multiplied by the frequency filter, with  $16$ random weights initialized close to $0.1$ and the fully-connected network weights initialized using Xavier method \cite{Glorot10understandingthe}. After this multiplication, the values inside each ring are aggregated by summing the frequencies inside it. Each value is then stacked and used as the input for the fully connected layer, containing a single hidden layer with $16$ \textit{ReLu} neurons. 
The output of this layer is then used as input for a \textit{softmax} layer with \textit{cross-entropy} loss, in order to give the probabilities for the \textit{cushion} and \textit{canvas} classes. 
The training process explored the stochastic gradient descent with momentum, considering learning rate $=0.01$, batch size $=4$, momentum $=0.9$ and epochs $=100$. 

\begin{figure}[!htbp]
 \centering
 \includegraphics[width=1\linewidth]{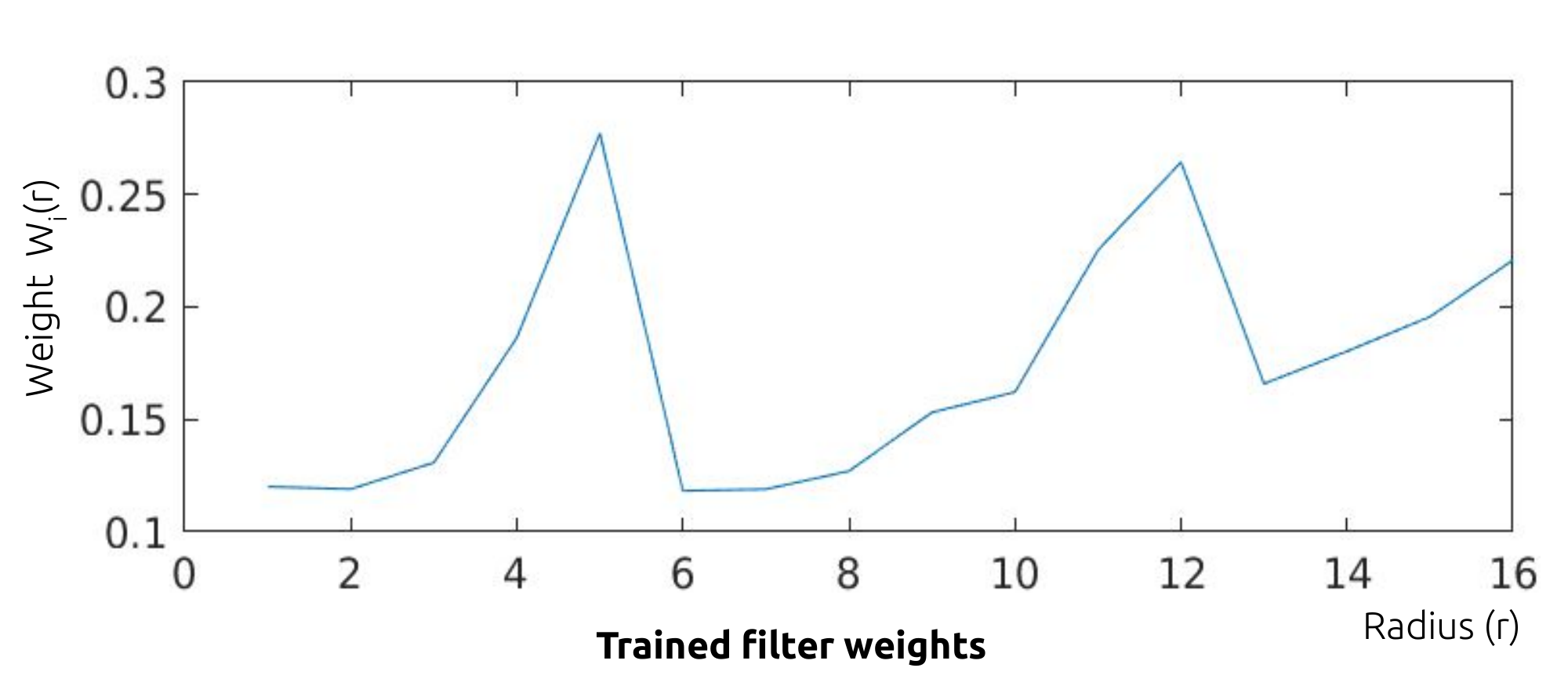}
 \caption{Trained frequency weights for \textit{canvas} and \textit{cushion} classification.}
 \label{fig:trained_canvas_cushion}
\end{figure}

\begin{figure}[!htbp]
 \centering
 \includegraphics[width=1\linewidth]{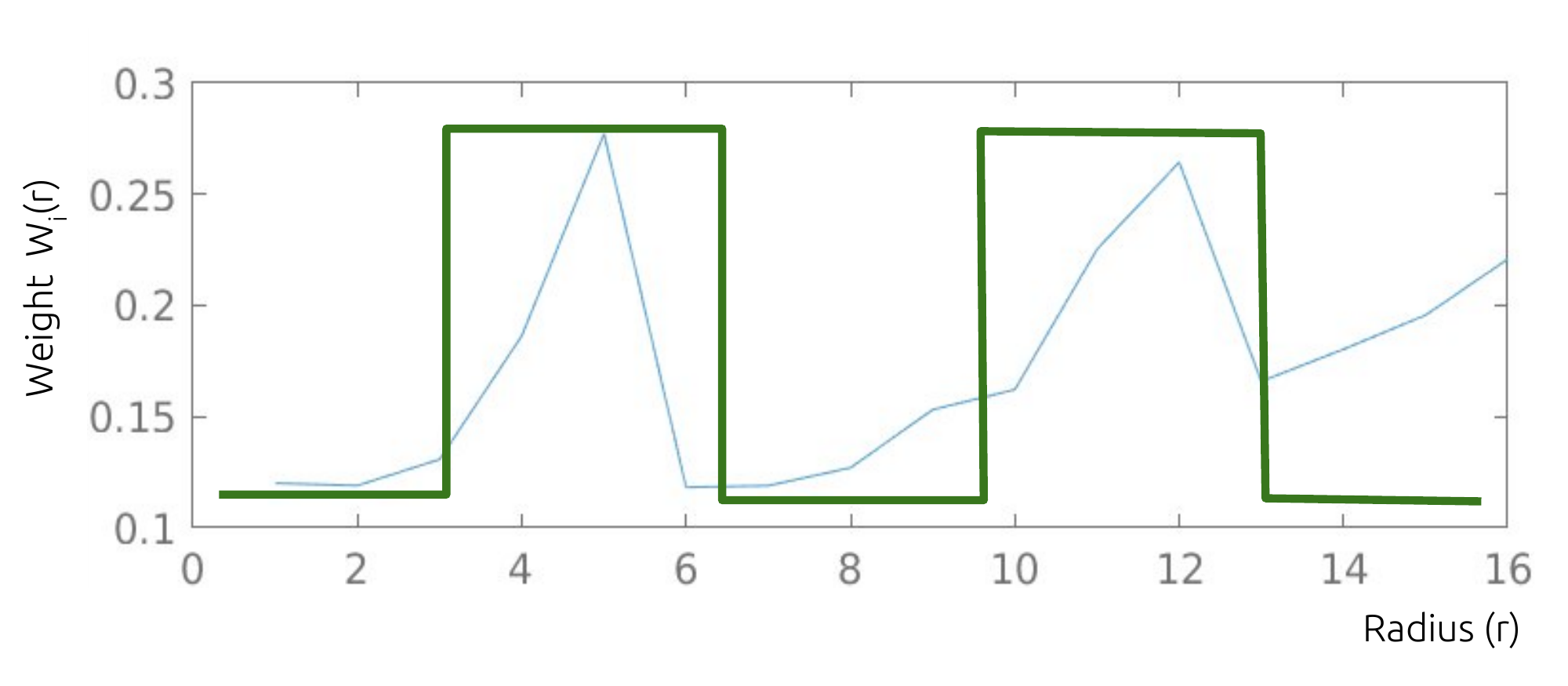}
 \caption{Trained filter weights overlapped with an ideal band-pass filter on distinguishable frequencies for \textit{canvas} and \textit{cushion}.}
 \label{fig:trained_canvas_cushion_band}
\end{figure}

Figure \ref{fig:trained_canvas_cushion} exhibits the obtained values for the filter weights as a function of the frequency radius.  As expected, the frequencies associated with radius $5$ and $12$ were emphasized by the filter, since these are the most discriminative components to separate the two classes. In this sense, the network created a band-pass filter that preserves the discriminative frequency components and cuts-off those that are not relevant for the classification, which is advantageous for this problem. In order to stress this aspect, we display in Figure \ref{fig:trained_canvas_cushion_band} an ideal band-pass filter, in green, along with the trained filter weights, in blue. 

\begin{figure*}[!htbp]
 \centering
 \includegraphics[width=1.0\linewidth]{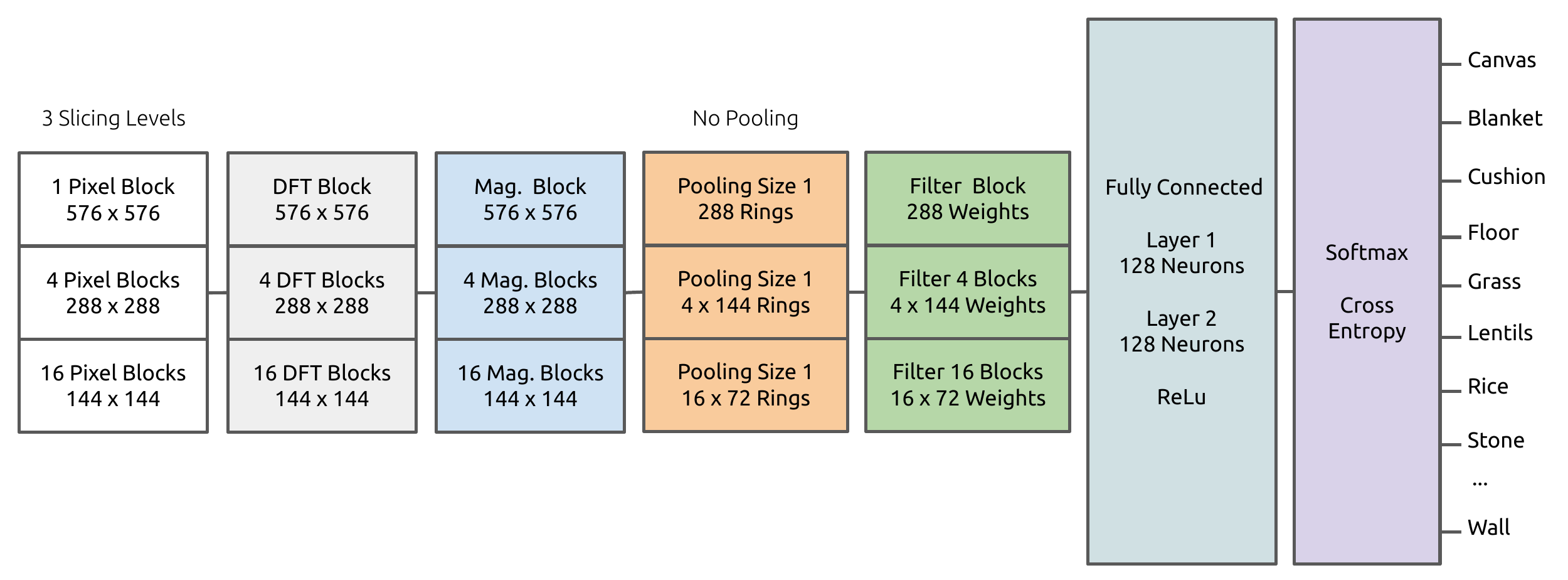}
 \caption{Architecture used to classify Kylberg 28 classes.}
 \label{fig:freq_net_28_classes}
\end{figure*}

After $100$ epochs of training, the accuracy in the validation set was $100\%$, which confirms that the proposed model adequately solved the classification task, and also that the learned frequency filter was capable of identifying the most discriminative frequency components in order to improve the classification.

Motivated by the successful results attained in this rather simple scenario, we proceed to more complex problems in order to evaluate the proposed method. 



\subsubsection{Experiment II - 28 textures classification}
\label{sec:28_textures_classification}

Now, we consider the whole Kylberg texture dataset. In this case, $28$ classes must be correctly classified. Figure \ref{fig:freq_net_28_classes} displays the architecture employed for this problem. 

The following parameter values were adopted for the training process: learning rate $=0.01$, learning rate decay $=0.005$, batch size $=1$, momentum $=0.9$ and epochs $=500$. We separated $75\%$ of the available images for the training set, while the remaining $25\%$ of the images were used in the validation set. 





The proposed model reached a validation accuracy of $99.73\%$, demonstrating a good capacity in separating several classes. 

\begin{figure}[!htbp]
 \centering
 \includegraphics[width=0.75\linewidth]{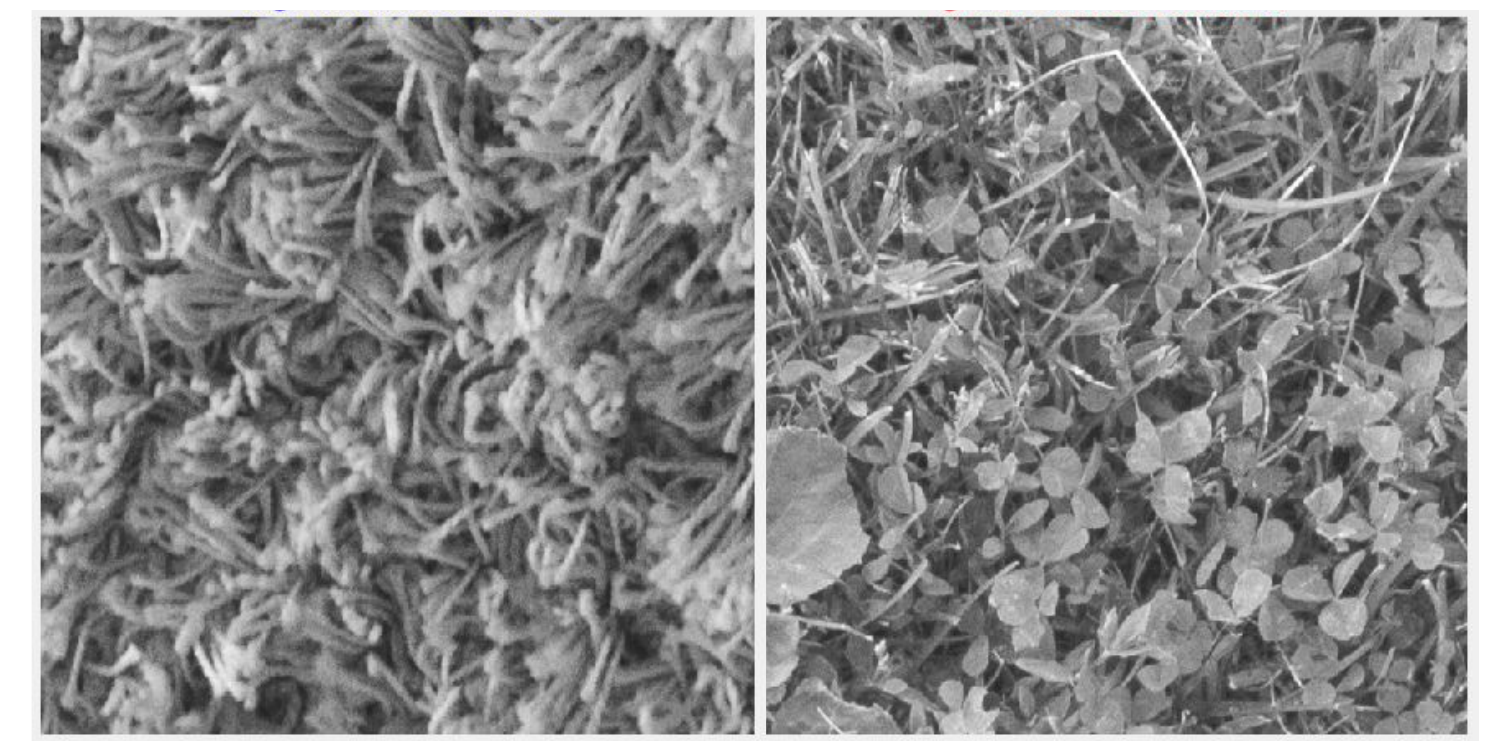}
 \caption{Classification error for 28 Kylberg textures: (\textit{rug} and \textit{grass} classes).}
 \label{fig:kylberg_28_classes_error}
\end{figure}

Figure \ref{fig:kylberg_28_classes_error} exhibits two examples of classification errors, in which the network confused the \textit{rug} and \textit{grass} classes. As we can notice, the images are, in fact, quite similar. 

It is important to stress that the majority of images from these two classes were correctly classified by the network, which means that this kind of misclassification is an exception, probably due to the internal variations in these two classes. 



\begin{figure}[!htbp]
 \centering
 \includegraphics[width=1\linewidth]{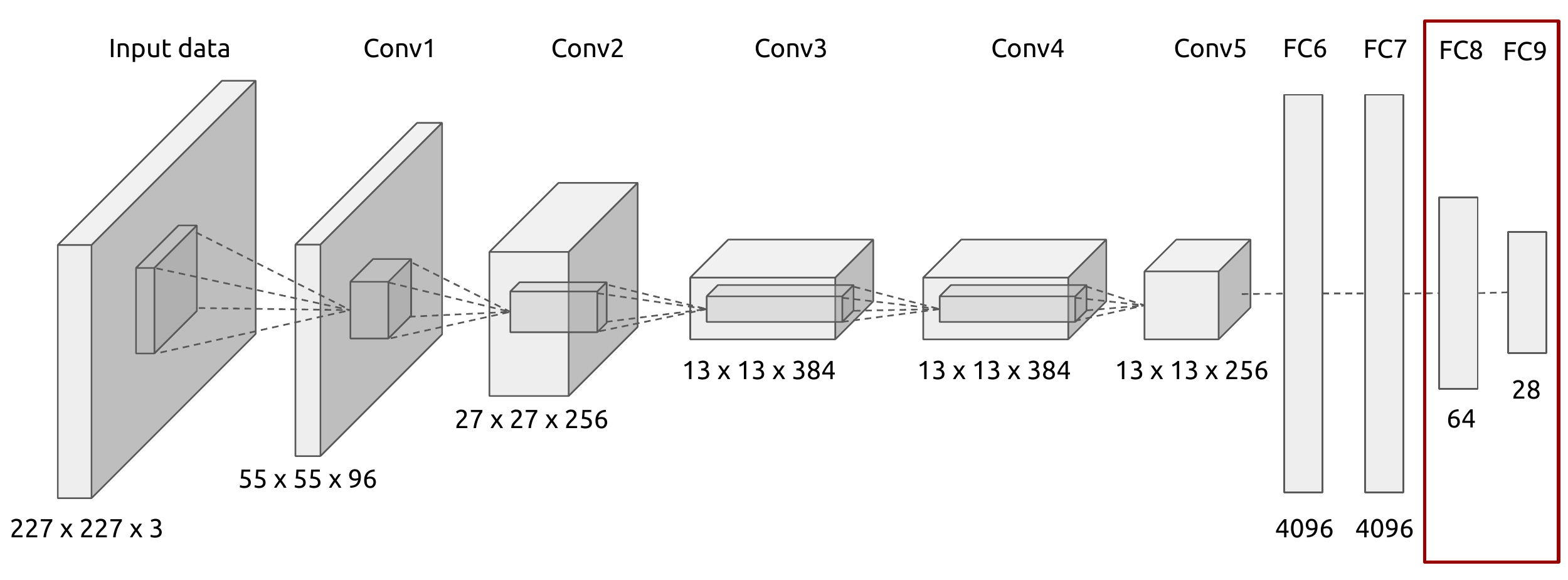}
 \caption{AlexNet architecture used for Kylberg dataset classification.}
 \label{fig:alexnet_28_classes}
\end{figure}

We also applied the renowned AlexNet architecture \cite{krizhevsky2012imagenet} in this problem, in order to analyze whether the proposed model is capable of attaining a performance similar to that of a deep convolutional network. In this case, we used transfer learning by retraining the structure of AlexNet pre-trained on ImageNet dataset \cite{imagenet_cvpr09}. The last layers of the architecture were modified in order to adapt it to the Kylberg texture classification problem, as shown in Figure \ref{fig:alexnet_28_classes}. 

Table \ref{tab:kylberg_tab} describes the characteristics (number of parameters and number of layers) of AlexNet and FreqNet, and presents the obtained classification accuracy in the validation set, as well as the computational time required for the training process. We tested different numbers of slicing levels, and the best results were obtained by using $3$ slices. 



\begin{table}[!htbp]
\centering
\caption{Comparison of deep learning architecture AlexNet and FreqNet for Kylberg texture classification.} 
\label{tab:kylberg_tab}
\begin{tabular}{@{\hspace{5pt}}c @{\hspace{6pt}}c@{\hspace{6pt}} cccc@{\hspace{5pt}}}
\hline
\multirow{1}{*}{\textbf{Method}} & \textbf{Parameters} & \textbf{Layers} & \textbf{Training time} & \textbf{Accuracy} \\
\hline
AlexNet & 61.00 million & 9 & 675m 47s & $\mathbf{99.82\%}$   \\
FreqNet & 0.276 million & 4 & 52m 36s & $99.73\%$   \\
\hline
\end{tabular}
\end{table}

\begin{figure*}[!htbp]
 \centering
 \includegraphics[width=1\linewidth]{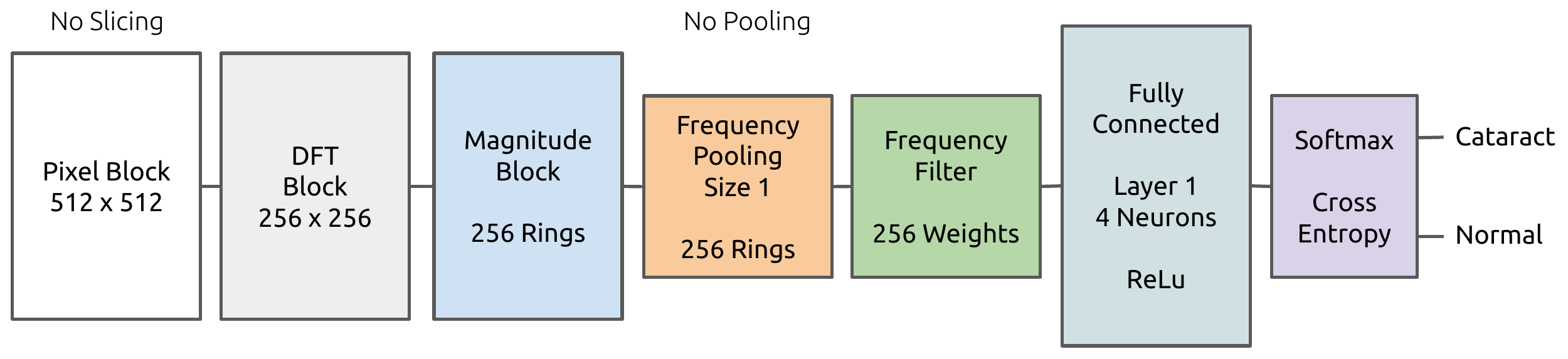}
 \caption{Architecture used to classify \textit{cataract} and \textit{normal} classes.}
 \label{fig:freq_net_cataract}
\end{figure*}

Although the deep learning architecture reached a superior accuracy when compared with FreqNet, it is possible to see that the number of adjustable parameters and the corresponding training time are much smaller in the FreqNet experiments. Additionally, the performances of both models were quite close, which indicates that the proposed method can be very competitive in this kind of classification task. 


\subsection{Retina Experiments}
\subsubsection{Cataract Dataset}

\begin{figure}[!htbp]
 \centering
 \includegraphics[width=1\linewidth]{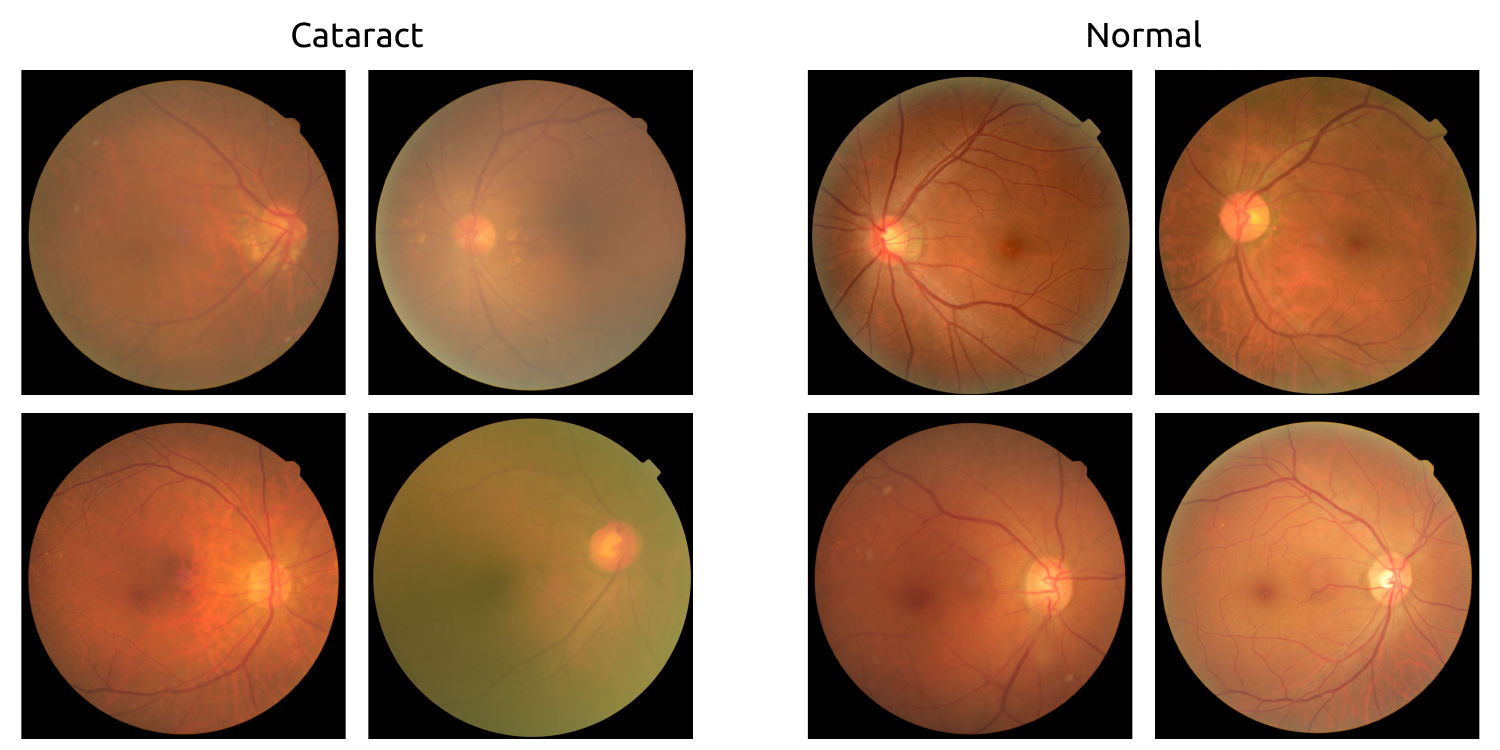}
 \caption{Images from Kaggle cataract dataset.}
 \label{fig:cataract_dataset}
\end{figure}

The next experiments involved medical datasets associated with retina images. Initially, we explored the Kaggle Cataract Dataset \cite{kaggle_cataract}, which is relatively small and contains $601$ retina images, being $300$ images of the \textit{normal} class, $100$ of the \textit{cataract} class, $101$ of the \textit{glaucoma} class and the remaining $100$ images are related to other retinal diseases, like diabetic retinopathy (DR). Since we are interested in evaluating the benefits of using frequency-domain information in classification, only the images from \textit{normal} and \textit{cataract} classes were employed in this experiment. Figure \ref{fig:cataract_dataset} presents some samples of this dataset: the two columns on the left present cataract images, whereas normal images are presented in the two columns on the right.  

It is possible to see that the cataract images are cloudy or blurred when compared to the normal eye. This happens since the crystalline becomes opaque in this disease and avoids the light to properly reach the retina. 

\subsubsection{Experiment III - Cataract detection}



The challenge in this experiment is to identify whether the patient has cataract or not by processing the retina image. In this case, the $400$ images comprising \textit{normal} and \textit{cataract} classes were separated in training ($75\%$) and validation ($25\%$) sets. 

Figure \ref{fig:freq_net_cataract}  presents the architecture of FreqNet explored in this problem. The training parameters were defined as follows: learning rate $=0.01$, learning rate decay $=0.1$, batch size $= 1$, momentum $=0.9$ and epochs $= 200$.



Analogously to the procedure explained in Section \ref{sec:28_textures_classification}, we also applied AlexNet to this problem using transfer learning and by adopting an output layer with two neurons, thus generating the probabilities of \textit{normal} and \textit{cataract} classes. The obtained results are summarized in Table \ref{tab:cataract_tab}.

\begin{table}[!htbp]
\centering
\caption{Comparison of deep learning architecture AlexNet and simple FreqNet for cataract classification based on retina images.} 
\label{tab:cataract_tab}
\begin{tabular}{@{\hspace{5pt}}c @{\hspace{6pt}}c@{\hspace{6pt}} cccc@{\hspace{5pt}}}
\hline
\multirow{1}{*}{\textbf{Method}} & \textbf{Parameters} & \textbf{Layers} & \textbf{Training time} & \textbf{Accuracy} \\
\hline
AlexNet & 61.00 million & 9 & 111m 35s & $87.00\%$   \\
FreqNet & 0.0012 million & 3 & 0m 17s & $\mathbf{93.00\%}$   \\
\hline
\end{tabular}
\end{table}

As we can observe, the FreqNet reached a superior performance when compared with AlexNet. Since this dataset is small, even using transfer learning for a pre-trained AlexNet model on ImageNet, the validation accuracy could not be improved, probably due to the number of parameters to be adjusted in this deep architecture. 

Another aspect that may have contributed to the success of FreqNet is that the frequency components for \textit{cataract} and \textit{normal} images usually are very different, since normal images have much more higher frequencies components than cataract images. So, by carefully designing the frequency filters, FreqNet can effectively explore these characteristics of the images to improve the classification.

Another interesting point to analyze refers to the number of parameters in both architectures. FreqNet is a very simple and lightweight model, with only $1.2$ thousand parameters, while AlexNet has more than $60$ million parameters. Still, the performance of the proposed method was much better, demonstrating that it can be an attractive solution for some classification problems. 

\begin{figure}[!htbp]
 \centering
 \includegraphics[width=1\linewidth]{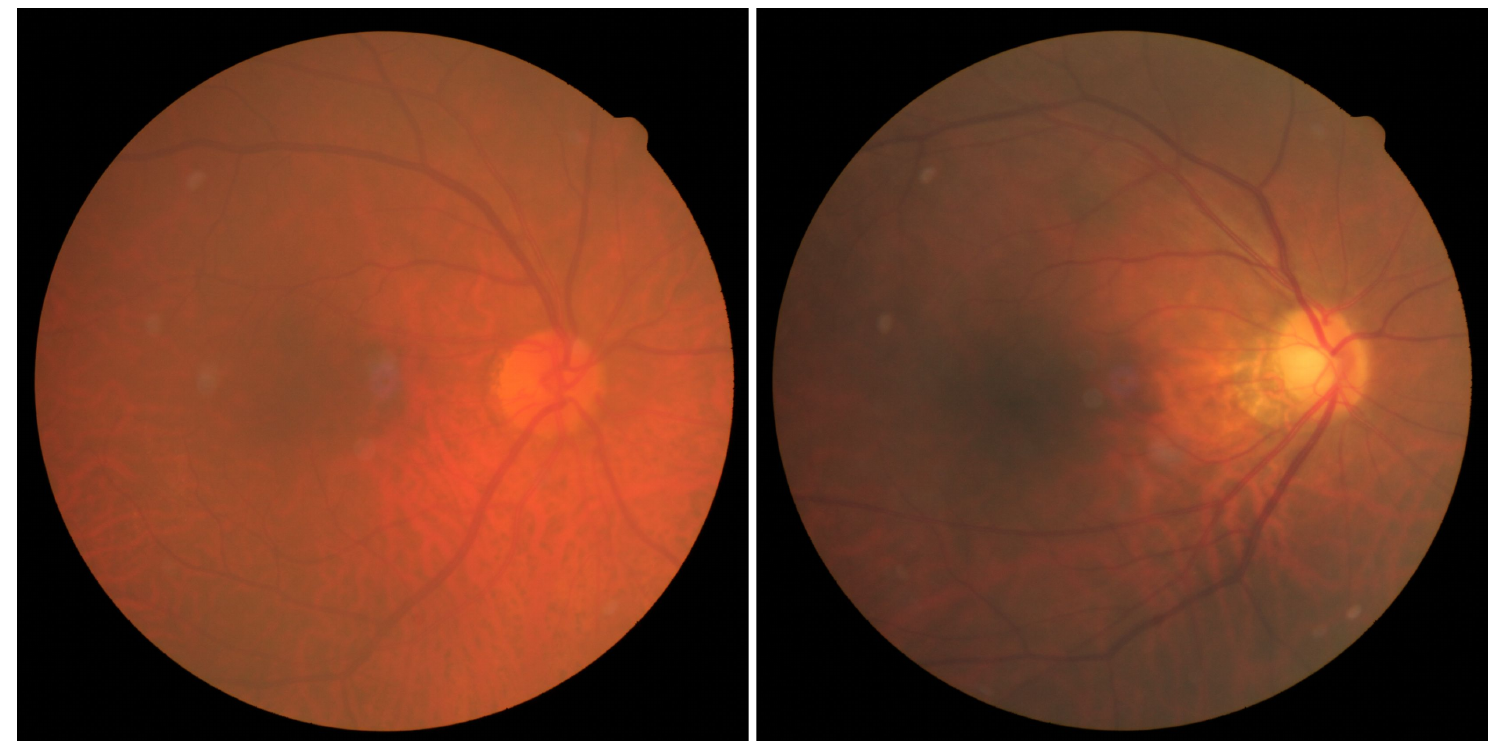}
 \caption{Normal retinas predicted as cataract.}
 \label{fig:cataract_2_classes_error}
\end{figure}

In Figure \ref{fig:cataract_2_classes_error}, we show two images of the \textit{normal} class that were classified as \textit{cataract} by the FreqNet. These classification errors may have occurred because the images are, to a certain extent, blurred in the same way as those in the \textit{cataract} class, but have been labeled as \textit{normal} by a human evaluator. 




In the next section a large retina dataset will be evaluated, but from the perspective of classifying the quality of the images. 

\subsubsection{EyePACS and EyeQ Dataset}

Assessing the retina image quality is an important factor in order to allow the correct diagnosis. Many images are captured with differences on illumination, focus and proper alignment with the eye fundus. The retina image quality index is also important to instruct the medical operator to repeat the image capture if it is out of focus, for example, or even can be used as a filter for machine learning algorithms that automatically analyze retina images in order to detect diseases like DR and glaucoma. An interesting dataset in this field is based on EyePACS project\cite{cuadros2009eyepacs}.

\begin{figure}[!htbp]
 \centering
 \includegraphics[width=1\linewidth]{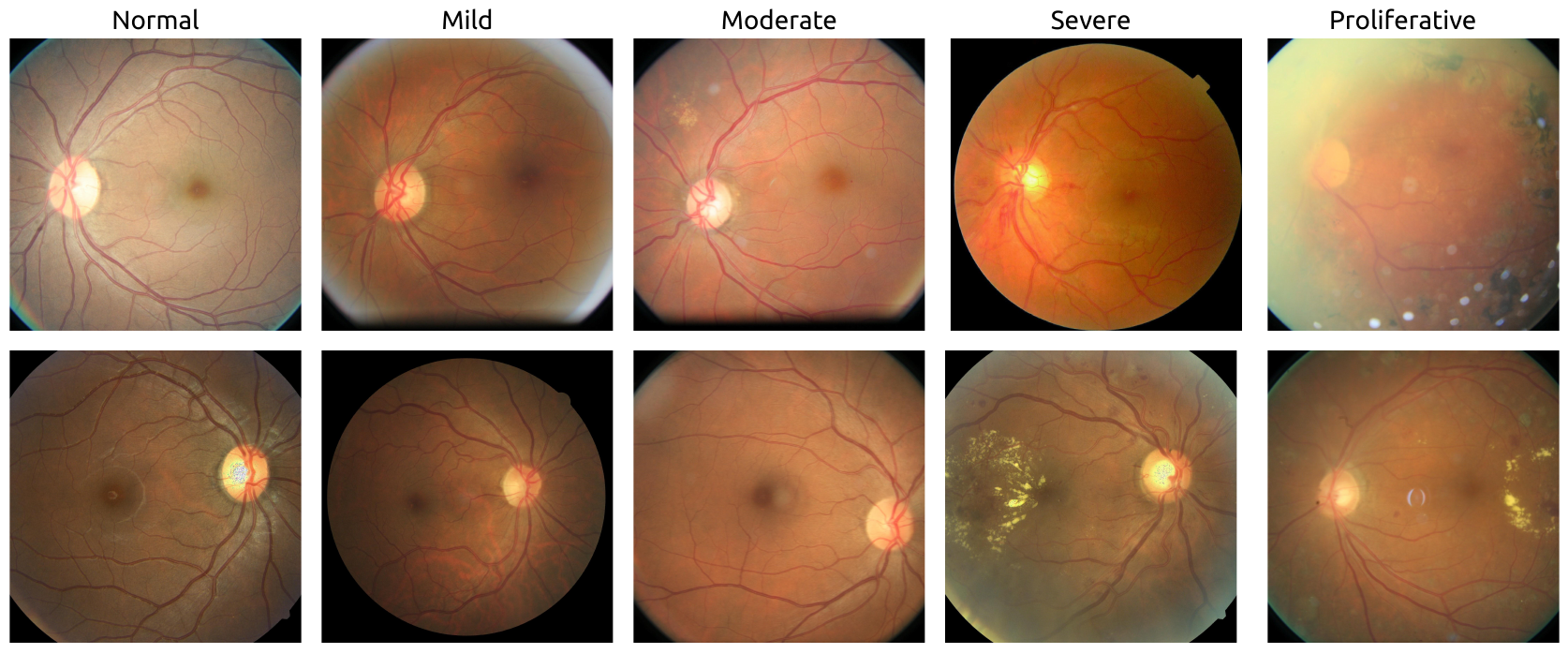}
 \caption{EyePACS dataset for diabetic retinophaty classification.}
 \label{fig:eyepacs_dataset}
\end{figure}

The EyePACS project provides an excellent dataset for DR detection and it is available on Kaggle platform\cite{kaggle_dr}. It provides about $90$ thousand retina images from $5$ different classes, each one identifying a DR level: \textit{normal}, \textit{mild}, \textit{moderate}, \textit{severe} and \textit{proliferative}. Figure \ref{fig:eyepacs_dataset}
presents examples of each class from this dataset.

This dataset is big and widely employed for DR classification, but it is a real world dataset, which means that, there is noise on images and labels. In this sense, the images contain artifacts being, for example, under or overexposed and also out of focus. 
This fact is quite pertinent for the current experiment, since we are interested in separating good from bad quality images.

\begin{figure}[!htbp]
 \centering
 \includegraphics[width=0.8\linewidth]{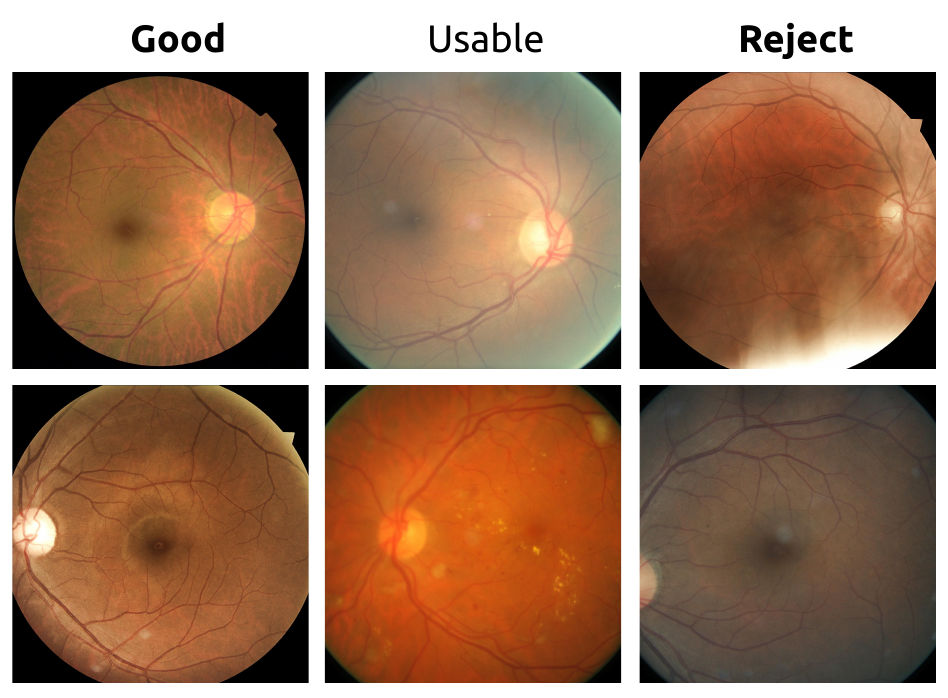}
 \caption{EyeQ dataset for retina quality evaluation.}
 \label{fig:eyeq_dataset}
\end{figure}

The EyeQ dataset\cite{eyeq} provides a subset of the EyePACS, dividing $28,792$ images in three classes according to the image quality: \textit{good}, \textit{usable} and \textit{reject}. Figure \ref{fig:eyeq_dataset} shows some examples from this dataset.

\subsubsection{Experiment IV - Retina Quality Evaluation}

\begin{figure*}[!htbp]
 \centering
 \includegraphics[width=0.98\linewidth]{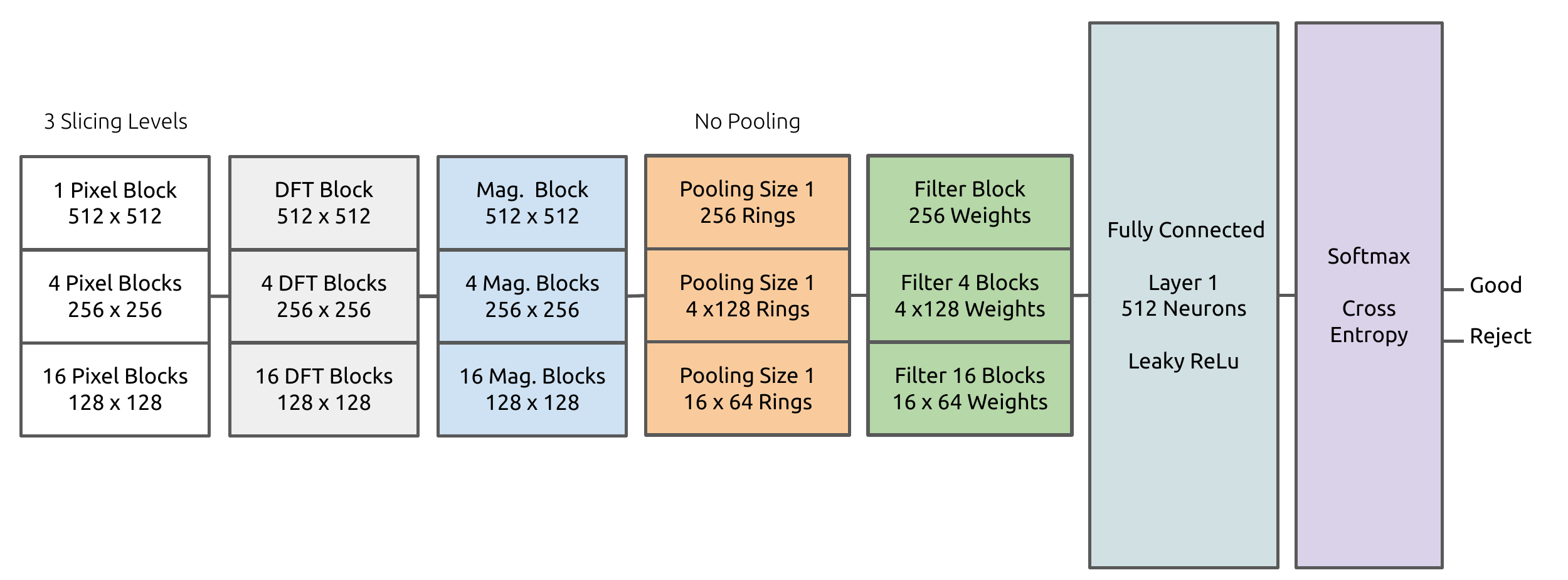}
 \caption{Architecture used to classify retina quality.}
 \label{fig:freq_net_retina_quality}
\end{figure*}


In this experiment, two classes of the EyeQ dataset were explored: \textit{good} and \textit{reject}. The idea is that by automatically recognizing the good and the bad images, it is possible to instruct an operator to repeat the acquisition whenever necessary, and, also, to help him/her during the examination. Considering these two classes, $22,358$ images were used for training and evaluating the models, according to the following division\cite{eyeq}: Training set -- $8,347$ images of \textit{good} category and $2,320$ images of \textit{reject} category; validation set -- $8,471$ images of \textit{good} category and $3,220$ images of \textit{reject} category. 


Using this dataset, we trained the FreqNet architecture exhibited in Figure \ref{fig:freq_net_retina_quality} using the following parameters: learning rate $=0.01$, learning rate decay $=0.1$, batch size $= 1$, momentum $= 0.9$ and epochs $= 1000$.

\begin{figure}[!htbp]
 \centering
 \includegraphics[width=0.75\linewidth]{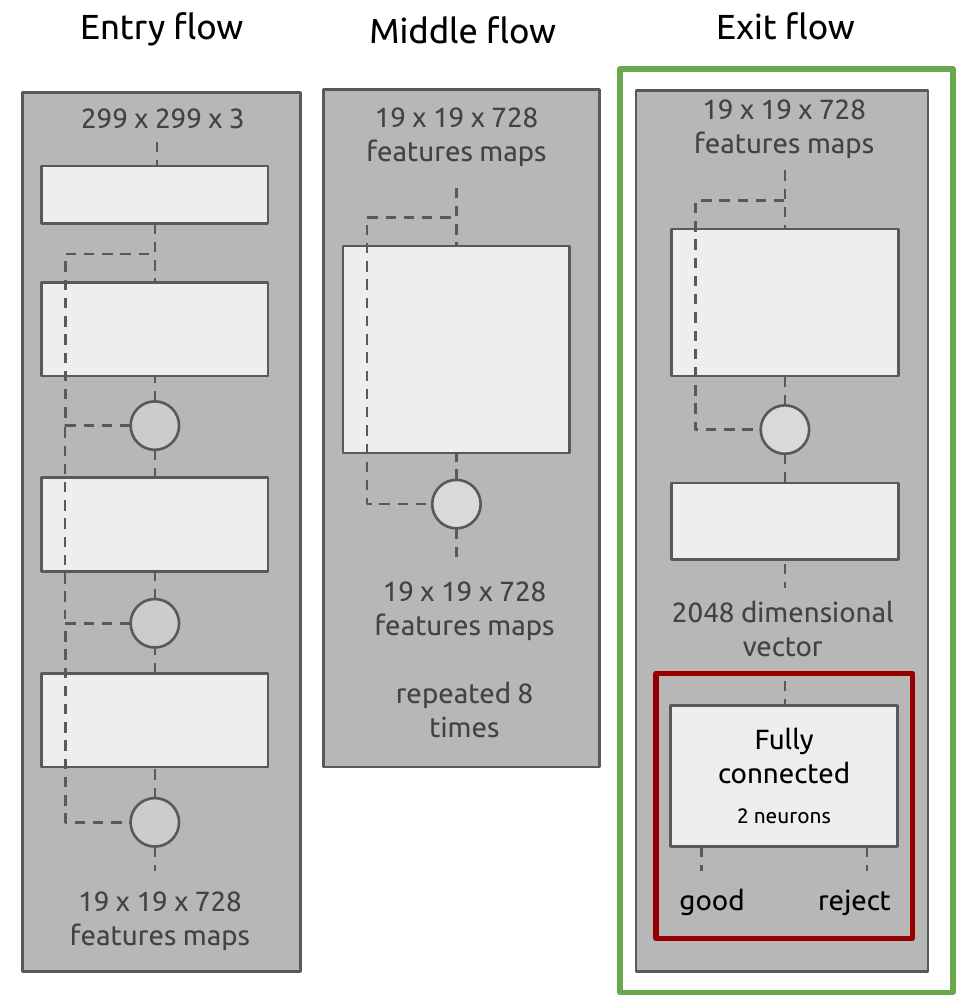}
 \caption{Xception architecture used for retina image quality evaluation (green box indicates the part of the network that was retrained).}
 \label{fig:xception}
\end{figure}

For the sake of comparison, AlexNet and Xception\cite{chollet2017xception} models were also trained. Both models were pre-trained on ImageNet dataset and transfer learning was applied in order to adapt them to the EyeQ images. 
AlexNet was trained following the same procedure mentioned in the previous experiments, considering two outputs. For the Xception architecture, only the exit flow was retrained using two units in the output layer. The modifications introduced in Xception are highlighted in Figure \ref{fig:xception}. 


The details concerning the architectures and the corresponding performances are presented in Table \ref{tab:quality_tab}. 

\begin{table}[!htbp]
\centering
\caption{Comparison of deep learning architecture AlexNet, Xception and FreqNet for retina quality classification.} 
\label{tab:quality_tab}
\begin{tabular}{@{\hspace{5pt}}c @{\hspace{6pt}}c@{\hspace{6pt}} cccc@{\hspace{5pt}}}
\hline
\multirow{1}{*}{\textbf{Method}} & \textbf{Parameters} & \textbf{Layers} & \textbf{Training time} & \textbf{Accuracy} \\
\hline
AlexNet & 61.00 million & 9 & 2242m 6s & $\mathbf{98.82\%}$   \\
Xception & 22.90 million & 71 & 19995m 49s & $98.08\%$   \\
FreqNet & 0.92 million & 3 & 762m 32s & $98.30\%$   \\
\hline
\end{tabular}
\end{table}

By analyzing the results, it is possible to observe that AlexNet reached the best accuracy, while FreqNet achieved the second best. One interesting point is that, again, the FreqNet is very competitive even with a much simpler and lightweight architecture, overcoming the Xception network. Another point for attention is the training time: it is possible to see that the FreqNet was trained in about $12$ hours; on other hand, the AlexNet training took about $1.5$ days, while the Xception training lasted almost $14$ days long. AlexNet and Xception were trained using Matlab Deep Learning toolbox implementation using a single GPU. On the other hand, FreqNet was trained in Matlab using only the CPU, which leaves a great opportunity for improvement. 

In fact, the FreqNet training time can be improved using more robust hardwares, which can also allow to train deeper frequency networks with more parameters, helping to improve even more the classification accuracy.
\section{Conclusion}
\label{sec:conc}


Frequency domain brings new possibilities to improve results in machine learning applications. Deep learning has achieved excellent results for many problems, but there is always room for improvement. Problems that have a great frequency dependency, like the ones related to image quality classification, may be suitably tackled by using simpler models based on the spectrum components.

Considering this, in this paper, a new method for training frequency domain layers for image classification was presented. The method applies the idea of iteratively splitting the image in small blocks, in order to extract global and local information from each part of the image. Each block is then transformed to the frequency domain by using the Discrete Fourier Transform. Using the magnitude of each block, frequency filters were trained based on the back-propagation algorithm.

The proposed method was analyzed in different datasets with the purpose of evaluating the performance on different problems. 
The first one was the Kylberg dataset, in order to distinguish $28$ classes of image textures. The results were very satisfactory, since it achieved $99.73\%$ of accuracy, while AlexNet achieved $99.82\%$. Despite the fact that the proposed method does not achieve the AlexNet performance, it is possible to see that it reached a competitive result using a very simple architecture, using $100\times$  less parameters than the AlexNet architecture.    

The second experiment involved a cataract dataset available in the Kaggle plataform. It is a small dataset, but allows to separate good retina from that ones with cataract in the eye. In this case, even using AlexNet pre-trained on ImageNet and retraining the entire network, the proposed method achieved a superior performance. In this case, the best result of AlexNet was $87.00\%$ of accuracy, while the proposed method achieved $93.00\%$. This probably happened since the normal and cataract retina images can be more easily distinguished in the frequency domain, since the cataract images are blurred and foggy due to the opaque crystalline caused by the disease. 

The third experiment was based on another retina dataset, but to distinguish good from bad images considering the image quality. In this case, three different architectures were employed: FreqNet, AlexNet and Xception architectures. This dataset was bigger and the AlexNet achieved the best performance with $98.82\%$, while FreqNet achieved $98.30\%$ and the Xception model $98.08\%$. Again, it is possible to notice that the proposed method achieved a competitive result with a much simpler model.

So, the main contributions of this work correspond to a novel architecture exploring the frequency domain, whose parameters are the frequency filter weights, that yielded competitive results when compared with DNNs architectures for some problems, but being much simpler and lightweight than these deeper architectures. In this sense, FreqNets are very attractive for mobile or real-time applications, since they have less parameters but with a satisfactory result, allowing to reduce, for example, the battery consumption when running in a mobile device or even allowing to process video, detecting objects or scenes in real-time.

The same idea proposed in this work can also be applied in other fields, for example, audio processing. In this context, MFCC features are used in many applications for years. However, they are the same for all the problems. In a simple example of classifying men and women voices, it is probably better to let the model train and boost the most discriminative frequencies that distinguish both genders, instead of using a generic feature extraction technique. Other applications on signal processing, like seismic signals detection, can also benefit from the proposed method. 


Finally, this work raises the discussion if the frequency domain can be mixed with spatial domain in order to increase the image classification performance. It is clear that some problems are better approached in frequency domain, like when approaching periodical signals. Exploring the traditional DNNs with frequency features can probably help to improve even more the results, being a very good field for research and future work.

\bibliographystyle{IEEEbib}
\bibliography{tpami}

\ifCLASSOPTIONcompsoc
\else
\fi






\begin{IEEEbiography}
[{\includegraphics[width=1in,height=1.25in,clip,keepaspectratio]{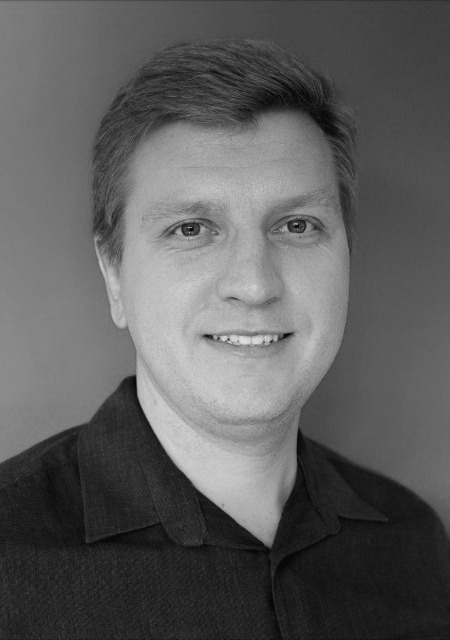}}]{Jos\'{e} Augusto Stuchi}
Received the degree in computer engineering (2009) and the MSc degree in electrical engineering (2013) from the University of São Paulo, and the MBA from Fundação Getúlio Vargas (2015). He is a PhD candidate on computer engineering at University of Campinas. He is the CEO of Phelcom Technologies, a Brazilian startup that develops medical devices in the eye care field, mixing optics, electronics, computer science and AI.

\end{IEEEbiography}

\begin{IEEEbiography}
[{\includegraphics[width=1in,height=1.25in,clip,keepaspectratio]{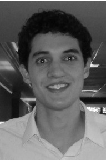}}]{Levy Boccato}
Received the titles of Computer Engineer (2008), Master in Electrical Engineering (2010) and Doctor in Electrical Engineering (2013), all from the University of Campinas (UNICAMP), S\~{a}o Paulo, Brazil. Currently, he is an Assistant Professor at the same university. His main research interests include computational intelligence, adaptive filtering, machine learning and signal processing.
\end{IEEEbiography}

\begin{IEEEbiography}
[{\includegraphics[width=1in,height=1.25in,clip,keepaspectratio]{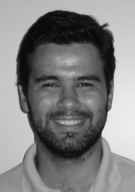}}]{Romis Attux}
Was born in Goiânia in 1978. He received the titles of Electrical Engineer (1999), Master in Electrical Engineering (2001) and Doctor in Electrical Engineering (2005) from the University of Campinas (UNICAMP). He is currently an associate professor at the same institution. His research interests are: adaptive signal processing, machine learning, brain-computer interfaces, chaotic systems and ethics in artificial intelligence.
\end{IEEEbiography}

\end{document}